\documentclass{article}

\PassOptionsToPackage{numbers, compress}{natbib}
\usepackage[preprint]{neurips_2024}
\usepackage{natbib}

\usepackage{subcaption}
\usepackage[utf8]{inputenc} % allow utf-8 input
\usepackage[T1]{fontenc}    % use 8-bit T1 fonts
\usepackage{url}            % simple URL typesetting
\usepackage{booktabs}       % professional-quality tables
\usepackage{amsfonts}       % blackboard math symbols
\usepackage{nicefrac}       % compact symbols for 1/2, etc.
\usepackage{microtype}      % microtypography
\usepackage{xcolor}         % colors

\usepackage{mathtools}
\usepackage[makeroom]{cancel}
\newcommand{\acro}[1]{\textsc{\MakeLowercase{#1}}}

\newcommand{\norm}[1]{\left\lVert#1\right\rVert}

\newcommand{\bftab}{\fontseries{b}\selectfont}
\DeclarePairedDelimiterX{\infdivx}[2]{[}{]}{%
  #1\;\delimsize\|\;#2%
}
\newcommand{\infdiv}{\text{KL}\infdivx}
\usepackage{amsmath}
\usepackage{amssymb}
\usepackage{amsthm}
\usepackage{float}
\usepackage{diagbox}
\usepackage{multirow}
\usepackage{siunitx}
\newcommand{\NA}{---}
\usepackage{nameref}
\usepackage{varioref}
\usepackage{hyperref}
\usepackage{cleveref}

\usepackage{tikz}
\usetikzlibrary{calc,positioning,shapes.geometric, fit}
\usetikzlibrary{3d,decorations.text,shapes.arrows,backgrounds}
\usetikzlibrary{matrix}

\newcommand{\drawTensor}[6]{% Arguments: T, N, K, xshift, yshift, label 1,2,3, scale
    \begin{scope}
        \draw[thick,fill=cyan!50!gray!50] (0,0) rectangle (#1,#2);
        \draw[thick, fill=green!20!gray!70] (#1,0)--(#1,#2)--++(2,0.4*#3)--++(0,-#2)--cycle;
        \draw[thick, fill=red!20] (0,#2)--(#1,#2)--++(2,0.4*#3)--++(-#1,0)--cycle;

        \foreach \i in {1,...,#1} {
            \draw[thin] (\i,0)--(\i,#2)--++(2,0.4*#3);
        }
        \foreach \i in {1,...,#2} {
            \draw[thin] (0,\i)--(#1,\i)--++(2,0.4*#3);
        }
       \foreach \i in {1,...,#3} {
            \draw[thin] ({2*\i/#3},{0.4*\i+#2})--++({#1},0)--++(0,-#2);
        }

        \node[below] at (#1/2,-0.1) {\Large #4};
        \node[above] at (#1/2+2,{#3*0.4+#2+0.1}) {\Large #5};
        \node[below right] at (#1+1+0.1,{#3*0.2+#2*0.5-0.3}) {\Large #6};
\end{scope}}

\newcommand{\drawMatrix}[6]{% Arguments: J, K, name, xshift, yshift, scale, label, block number
    \begin{scope}
        \draw[thick,fill=cyan!10!gray!10] (0,0) rectangle (#1,#2);

        \ifnum#6=1
            \pgfmathsetseed{13407} 
            \ifx#1#2
                \foreach \i in {1,...,#1} {
                    \foreach \j in {\i,...,#2} {
                        \pgfmathparse{rnd}
                        \ifdim \pgfmathresult pt > 0.75pt
                        \draw[fill=cyan!80!gray!80] (#1-\i,\j-1) rectangle ++(1,1);
                        \draw[fill=cyan!80!gray!80] (#1-\j,\i-1 ) rectangle ++(1,1);
                        \fi
                    }
                }
            \else
                \foreach \i in {1,...,#1} {
                    \foreach \j in {1,...,#2} {
                        \pgfmathparse{rnd}
                        \ifdim \pgfmathresult pt > 0.75pt
                         \draw[fill=cyan!80!gray!80] (\i-1,\j-1) rectangle ++(1,1);
                        \fi
                    }
                }
            \fi
        \else
            \foreach \i in {1,...,#6} {
                 \fill[cyan!80!gray!80] ({#1-\i*#1/#6},{\i*#2/#6-#2/#6}) rectangle ++({#1/#6},{#2/#6});
            }
        \fi
        
        \foreach \i in {1,...,#1} {
            \draw[thin] (\i,0)--(\i,#2);
        }
        \foreach \i in {1,...,#2} {
            \draw[thin] (0,\i)--(#1,\i);
        }

        \node[below] at (#1/2,-0.1) {\Large #4};
        \node[left] at (-0.1,#2/2) {\Large #3};
        \node[above] at (#1/2,#2+0.1) {\Large #5};
\end{scope}}

\title{Idiographic Personality Gaussian Process for Psychological Assessment}

\author{%
  Yehu Chen, Muchen Xi, Jacob Montgomery\\ \textbf{Joshua Jackson, Roman Garnett} \\
  Washington University in St Louis\\
\texttt{chenyehu,m.xi,j.jackson,jacob.montgomery,garnett@wustl.edu}
}

\begin{document}

\maketitle

\begin{abstract}
% Developing taxonomies for psychological assessment is crucial for understanding long-term human behaviors. However, existing psychometric methods tend to be nomothetic, lacking individualization, or rely on static cross-sectional data that overlooks the dynamic nature of psychological processes. We introduce an idiographic personality Gaussian process (\acro{IPGP}) framework for time-series survey data, by leveraging Gaussian process coregionalization to conceptualize individualized taxonomies and stochastic variational inference for computational scalability. Through an extensive simulation study against benchmark methods and an exploratory factor analysis study of life outcomes of personality replication, we demonstrate that \acro{IPGP} can simultaneously improve estimation of idiographic taxonomies and prediction of missing responses. We also assess \acro{IPGP} using our IRB-approved data with a forecasting and a leave-one-trait-out prediction task, illustrating how \acro{IPGP} identifies unique taxonomies of personality that display potential in advancing individualized approaches to psychological diagnosis.
We develop a novel measurement framework based on a Gaussian process coregionalization model to address a long-lasting debate in psychometrics: whether psychological features like personality share a common structure across the population, vary uniquely for individuals, or some combination. We propose the idiographic personality Gaussian process (\acro{IPGP}) framework, an intermediate model that accommodates both shared trait structure across a population and ``idiographic'' deviations for individuals.  
\acro{IPGP} leverages the Gaussian process coregionalization model to handle the grouped nature of battery responses, but adjusted to non-Gaussian ordinal data. We further exploit stochastic variational inference for efficient latent factor estimation required for idiographic modeling at scale.
Using synthetic and real data, we show that \acro{IPGP} improves both prediction of actual responses and estimation of individualized factor structures relative to existing benchmarks. In a third study, we show that \acro{IPGP} also identifies unique clusters of personality taxonomies in real-world data, displaying great potential in advancing individualized approaches to psychological diagnosis and treatment. 
\end{abstract}

\section{Introduction}\label{sec:intro}

Building models for assessment of latent traits from observed responses is crucial to understand long-term behaviors through repeated quantitative assessments.  These are used, for instance, to study emotional stability after medical treatment, or development of academic ability during secondary education \citep{molenaar2004manifesto, wang2013bayesian, dumas2020dynamic}. However, existing frameworks face several interrelated limitations. First, there are strong reasons to believe that standard taxonomies may over-generalize, failing to distinguish between related psychological phenomenon that often differ in etiology, symptoms, and biological processes across individuals \citep[e.g.,][]{molenaar2004manifesto, borsboom2003theoretical}. A related issue is that measurement models are rarely individualized, instead assuming that (1) the correlation \emph{between} latent traits of interest and survey responses are invariant across individuals and (2) the relationship between the latent trait and the observed quantitative indicators are the same for everyone. Lastly, current models are almost always developed for cross-sectional data that are collected only once from each respondent, which overlooks the dynamics of any psychological process.

To address these limitations, previous research has adopted three different approaches, each inadequate in its own way.  First, recent work has proposed an \textit{idiographic} approach that builds a completely distinct taxonomy for everyone \citep{borkenau1998big,beck2020consistency, beck2021within}. However, complete personalization may sacrifice generalizability and interpretability to clinicians, since any possible population commonality is completely ignored. A second line of research focuses on building dynamic psychometric models of time-series data via some variant of item response theory \citep{rijmen2003nonlinear, reise2009item, dumas2020dynamic}, longitudinal structural equation modeling \citep{little2013longitudinal, kim2014testing, asparouhov2018dynamic}, vectorized autoregression \citep{lu2018bayesian, haslbeck2021tutorial} and/or Gaussian process (GP) latent trajectories \citep{NIPS2005_ccd45007, damianou2011variational, durichen2014multitask}. Yet all these models adopt the \textit{nomothetic} approach, assuming that responses from all individuals share an identical latent structure. Finally, there is a smaller body of work that adopts an intermediate approaches for creating individualization while maintaining group commonality \citep[e.g.,][]{beltz2016bridging}. However, prior research models quantitative responses directly, ignoring the latent structures that are the actual focus of domain researchers.  

In this work, we propose an idiographic personality Gaussian process (\acro{IPGP}) framework for assessing dynamic psychological taxonomies from time-series survey data, and combine the nomethetic and idiographic approaches by deploying a common structure for explaining the typical circumstance and individual structures for permitting deviations into distinct forms. We leverage the Gaussian process coregionalization model to conceptualize responses of grouped survey batteries, adjusted to non-Gaussian ordinal data, and utilize \acro{IPGP} for hypothesis testing of domain theories. Computationally, our framework also exploits the stochastic variational inference for latent factor estimation, contrasting with other \acro{GP} measurement models relying on Gibbs sampling that may not scale efficiently to intensive longitudinal setups \citep{durichen2014multitask, duck2020gpirt}.

To our knowledge, our work is the first multi-task Gaussian process latent variable model for dynamic idiographic assessment. Existing models either focus on cross-sectional settings (with no dynamics)  \citep{borkenau1998big, bonilla2007multi, beck2021within} or single-task settings where there is no inter-battery correlation \citep{Snelson_2005, hensman2015scalable}. Methodologically, our approach advances the literatures on Gaussian process latent variable models (\acro{GPLVM}) \citep{lawrence2003gaussian}, Gaussian process dynamic systems (\acro{GPDM}) \citep{damianou2011variational, durichen2014multitask} and \acro{GP} ordinal regression for likert-type survey data \citep{croasmun2011using, JMLR:v6:chu05a}. Through an extensive simulation study against benchmark methods and an  analysis of an existing cross-sectional personality data, we demonstrate that \acro{IPGP} can simultaneously improve estimation of idiographic taxonomies and prediction of missing responses. We also assess \acro{IPGP} using an IRB-approved longitudinal pilot study.  We show  \acro{IPGP} offers superior performance in predicting responses and and illustrate how \acro{IPGP} identifies unique taxonomies of personality that display potential in advancing individualized approaches to psychological diagnosis and treatment.

\section{Backgrounds}\label{sec:background}
We start by laying out the ordinal factor model for building standard taxonomy from survey data \citep{digman1997higher, baglin2014improving}. We then briefly discuss several existing idiographic longitudinal models in psychological assessment, and review the Gaussian process model. 

\paragraph{Ordinal factor analysis.} Consider the scenario where some set of units, $i\in \{1,\dots,N\}$, repeatedly answering the same set of $j\in \{i,\dots,J\}$ survey items over $t\in \{1,\dots,T\}$ periods with ordinal observations $y_{ijt}\in \{1,\dots,C\}$ up to $C$ levels. For example, the responses could be Likert-typed, ranging from ``strongly disagree'' to ``strongly agree''. The latent factor model posits that the $j$th underlying latent variable $f_{j}^{(i)}(t)$ for unit $i$ at time $t$ are factored as $\mathbf{w}^T_{j}\mathbf{x}_{i}(t) $, where $\mathbf{x}_{i}(t)\in\mathbf{R}^K$ are unit-level latent factors  and $\mathbf{w}_{j}\in\mathbf{R}^K$ are factor loadings. The $f_{j}^{(i)}(t)$s are then mapped to ordinal responses via an ordered logit model: $ p\big(y_{ijt}=c\mid f_{j}^{(i)}(t)=f\big) = \Phi(b_{c}-f) - \Phi(b_{c-1}-f)$ with threshold parameters $b_0<\dots<b_C$. Usually $b_0$ and $b_C$ are fixed to $-\infty$ and $+\infty$ such that the resulted categorical probability vector sums to $1$, while $b_1,\dots,b_{C-1}$ are allowed to move freely. Stacking $\mathbf{x}_{i}(t)$s, $\mathbf{w}_{j}$s and $y_{ijt}$'s into matrices $\mathbf{x}$, $\mathbf{w}$ and tensor $\mathbf{y}$, the joint likelihood can be written as $\mathcal{L}(\mathbf{y}\mid \mathbf{x},\mathbf{w})=\prod_{i}\prod_{j}\prod_t{p}\big(y_{ijt}\mid \mathbf{x}_{i}(t), \mathbf{w}_{j}\big)$, while model identification is guaranteed by the general rule of factor models with additional orthogonality and normalization constraints \citep{bollen1989structural}. This factor model is also known as item response model \citep{samejima1969estimation,van1997handbook}, which estimates parameters via maximum likelihood, weighted least squares, or an \acro{EM} algorithm \citep{bock1981marginal, Forero2009factor, li2016confirmatory}.

\paragraph{Idiographic longitudinal assessment.} In psychological assessment, the idiographic approach emphasizes \textit{intrapersonal} variation by requiring distinct loadings $\mathbf{w}^{(i)}_{j}$, while the nomothetic approach identifies general \textit{interpersonal} variation assuming shared factor loadings $\mathbf{w}_{j}$ \citep{salvatore2010between}. In terms of data collection, the idiographic approach usually surveys each individual multiple times ($n=1$ and large $T$) for learning personalized taxonomy rather than many individuals at a single shot (large $n$ and $T=1$). To extract individualized dynamics from time-series data, recent psychometric models have utilized longitudinal structural equations by explicitly specifying any intrapersonal and temporal dynamics. However, these typically require strong model-based assumptions from domain-theory about this dynamic process, and may be sensitive to model mis-specification \citep{little2013longitudinal, asparouhov2018dynamic}. Meanwhile, 
variants of hierarchical vector autoregression may automatically learn individual trajectories over time, but are designed to model observed responses directly rather than the latent traits of interest to domain scholars \citep{lu2018bayesian, haslbeck2021tutorial}.

\paragraph{Gaussian process.}  A Gaussian process (\acro{GP}) can be used to define a distribution over $f$ such that the evaluation of $f$ at arbitrary subset of $\mathcal{X}$ is a joint multivariate Gaussian \citep{rasmussen2006gpml}. To determine its mean and covariance, a $\mathcal{GP}(\mu, K)$ is specified with a mean function $\mu: \mathcal{X} \rightarrow \mathbf{R}$ and a positive-definite kernel function $K: \mathcal{X}\times \mathcal{X} \rightarrow \mathbf{R}$. The most common kernel is the squared exponential (\acro{RBF}) kernel $K(\mathbf{x_1},\mathbf{x_2})=\exp{(-\frac{1}{2}\mathbf{x_1}^T\mathbf{P}\mathbf{x_2}})$ with precision matrix $\mathbf{P}=\mathbf{diag}(1/\ell^2_1,\dots, 1/\ell^2_d)$ and $d=\mathbf{card}(\mathcal{X})$. Posterior of a \acro{GP} is usually analytical for Gaussian likelihood, but needs to be approximated in modeling latent variables. We discuss the variational approximation in Sec. (\ref{sec:methodology}).

\begin{figure*}[t!]
    \centering 
    \resizebox{\textwidth}{!}{
\begin{tikzpicture}[scale=1]
\node at (0,0) (Wpop) {\begin{tikzpicture}[scale=0.5]
\drawMatrix{6}{3}{$K$}{$J$}{$\mathbf{W}_{\text{pop}}$}{3};
\end{tikzpicture}};

\node[right=0.5 of Wpop] (Kpop) {\begin{tikzpicture}[scale=0.4]
\drawMatrix{6}{6}{}{}{$\mathbf{K}_{\text{pop}}$}{3};
\end{tikzpicture}};

\node[right=0.1 of Wpop] (Wpoparrow) {\LARGE $\Rightarrow$};

\node[below=1 of Wpop] (Wi) {\begin{tikzpicture}[scale=0.5]
\drawMatrix{6}{3}{$K$}{$J$}{$\mathbf{w}_{i}$}{1};
\end{tikzpicture}};

\node[right=0.5 of Wi] (Ki) {\begin{tikzpicture}[scale=0.4]
\drawMatrix{6}{6}{}{}{$\mathbf{K}_{\text{ind}}$}{1};
\end{tikzpicture}};
\node[right=0.1 of Wi] (Wiarrow) {\LARGE $\Rightarrow$};

\node (loading) [draw=red,rounded corners, dashed, fit = (Wpop) (Wi), inner xsep=2pt] {\\
\large \textsf{Factor Loadings}};

\node (loading) [draw=red,rounded corners, dashed, fit = (Kpop) (Ki), inner sep=1pt] {\\
\large \textsf{Cov Matrix}};

\node[below right=-3.3 and 1.6 of Ki] (Ktime) {\begin{tikzpicture}[scale=0.4]
\drawMatrix{4}{4}{}{}{$\mathbf{K}^{(i)}_{\text{time}}\sim \mathbf{RBF}$}{2};
\end{tikzpicture}};

\node[above=0.4 of Ktime] (xi) {\begin{tikzpicture}[scale=0.5]
\drawTensor{4}{1}{3}{$T$}{\large \textsf{Factors}  $\mathbf{x}_i(t)$}{$K$};
\end{tikzpicture}};
\node[above=0.0 of Ktime] (xiarrow) {\LARGE $\Uparrow$};

\node[right=6 of Ki] (y) {\begin{tikzpicture}[scale=0.4]
\drawTensor{4}{1}{6}{$T$}{ $\mathbf{y}^{(i)}$}{$J$};
\end{tikzpicture}};
\node (factor) [draw=red,rounded corners, dashed, fit = (xi) (Ktime) (xiarrow), inner sep=0] {};

\node[above=0.5 of xi] (product) {\LARGE $\otimes$};

\node[above left=0.4 and 0.6 of Ktime] (plus) {\LARGE $\oplus$};
\draw[->,line width=1.5pt, to path={-| (\tikztotarget)}] (Kpop) edge (plus);
\draw[->, line width=1.5pt,to path={-| (\tikztotarget)}] (Ki) edge (plus);
\draw[->,line width=1.5pt] (plus) to ++(0.5,0) to[to path={|- (\tikztotarget)}] (product);
\draw[->,line width=1.5pt] (xi) edge (product);

\node[above=-0.4 of y] (yarrow) {\LARGE $\Downarrow$};
% \node[above=0.5 of y] (lik) {\large \textsf{Ordinal likelihood}};

\node[above=0.9 of y] (farrow) {\LARGE $\Downarrow$};
\node[above=1.1 of y] (f) {\begin{tikzpicture}[scale=0.4]
\drawTensor{4}{1}{6}{$T$}{ $\mathbf{f}^{(i)}$}{$J$};
\end{tikzpicture}};

\node (lik) [draw=red, rounded corners, dashed, fit = (f) (y) (farrow) (yarrow), inner xsep=8pt, inner ysep=1pt] { \large \textsf{Ordinal Model}};

\draw[->,line width=1.5pt] (product) to ++(2.2,0) to[to path={|- (\tikztotarget)}] (f);
\end{tikzpicture}}
    \caption{{\small Proposed \acro{IPGP} model for inferring latent factors and factor loadings from dynamic ordinal data. Input ordinal observations across channels are modeled as ordinal transformations of latent dynamic Gaussian processes with individualized \acro{RBF} kernels and loading matrices.}}
    \label{fig:architecture}
\end{figure*}
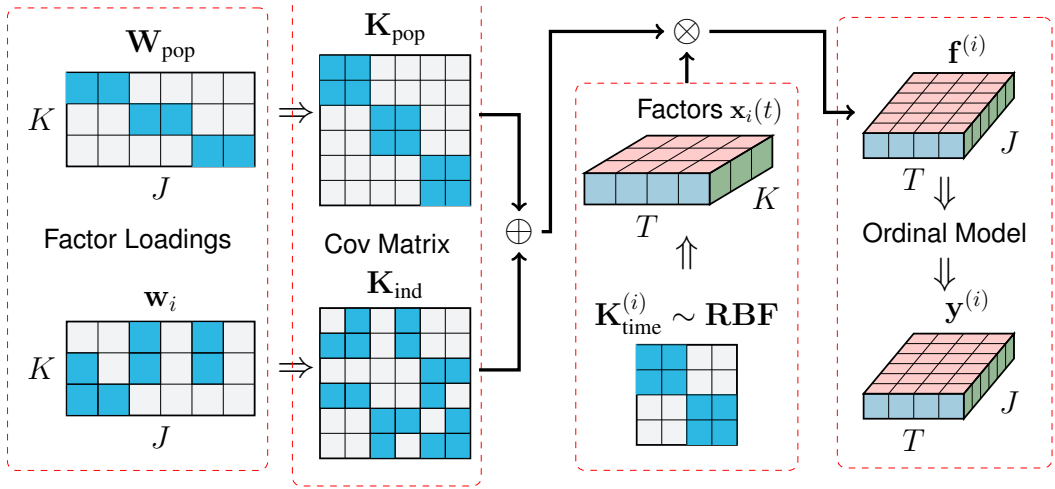

\section{Methodology}\label{sec:methodology}
% In this regard, our approach resembles the probability principle component analysis \citep{lawrence2003gaussian} but targets the non-Gaussian ordinal likelihood through stochastic variational inference with promising scalability. 
% for its faster convergence than Markov Chain Monte Carlo methods while not satisfying inference accuracy \citep{girolami2006variational, bonilla2019generic}

We propose an idiographic personality Gaussian process (\acro{IPGP}) framework for assessing individualized dynamic psychological taxonomies from time-series survey data. Instead of joint estimation of latent factors and their loadings that cannot guarantee rotational and scaling invariance, we marginalize out the latent variables and focus on learning taxonomies of loadings. The overall architecture of \acro{IPGP} is illustrated in Figure (\ref{fig:architecture}), where input ordinal observations across channels are modeled as ordinal transformations of latent dynamic \acro{GP} with individualized \acro{RBF} kernels and loading matrices.

%   We adopt variational inference for its faster convergence than Markov chain Monte Carlo methods while not satisfying accuracy \citep{girolami2006variational, bonilla2019generic}.
% Finally, our framework permits hypothesis testing to address the substantive question of nomothetic versus idiographic approach via Bayesian model comparison.

\subsection{Multi-task learning}

Typically in psychological assessment, survey questions are meticulously grouped such that each group gauges a particular latent trait (e.g., facet of personality). Hence, we conceptualize the assessment of psychological traits as a multi-task learning problem, where each question represents a distinct task but can be correlated with other tasks. A multi-task \acro{GP} is an extension of the single-task \acro{GP} but for vector-valued functions \citep{bonilla2007multi}. To motivate the multi-task framework, first consider the two-task scenario with two $T\times 1$ vector $\mathbf{f_{1}}^{(i)}$ and $\mathbf{f_{2}}^{(i)}$ denoting the latent temporal processes of unit $i$ for question $j=1,2$. To fix the scale of latent factors, a time-level Gaussian process prior is placed on $\mathbf{x}_{i}(t)\sim\mathcal{GP}(\mathbf{0},\mathbf{K}^{(i)}_{\text{time}})$. Hence, by exploiting affine property of Gaussian, the induced joint distribution of vectorized $[\mathbf{f_{1}}^{(i)},\mathbf{f_{2}}^{(i)}]^T$ can be written as:
\begin{equation}
\begin{bmatrix}
\mathbf{f_{1}}^{(i)} \\
\mathbf{f_{2}}^{(i)}
\end{bmatrix}
    \sim \mathcal{GP}\Big(\begin{bmatrix}
\mathbf{0} \\
\mathbf{0}
\end{bmatrix},
\begin{bmatrix}
\mathbf{w}^T_{1}  \mathbf{w}_{1} \mathbf{K}^{(i)}_{\text{time}} & \mathbf{w}^T_{1}  \mathbf{w}_{2} \mathbf{K}^{(i)}_{\text{time}} \\
\mathbf{w}^T_{2}  \mathbf{w}_{1} \mathbf{K}^{(i)}_{\text{time}} & \mathbf{w}^T_{2}  \mathbf{w}_{2} \mathbf{K}^{(i)}_{\text{time}}
\end{bmatrix} \Big)
\end{equation}
whose covariance of shape $2T\times 2T$ contains four block matrices $\mathbf{K}^{(i)}_{\text{time}}$ scaled by different $\mathbf{w}^T_{j}\mathbf{w}_{j'}$ ($j,j'\in \{1,2\}$). Specifically, $\mathbf{w}^T_{1}\mathbf{w}_{2}$ controls the inter-task covariance between these two tasks and $\mathbf{w}^T_{j}\mathbf{w}_{j}$s ($j\in \{1,2\}$) control their intra-task variance. This multi-task structure is also known as the linear model of coregionalization (\acro{LMC}) \citep{alvarez2012kernels}, where the factor structure can be recovered from the relations $\mathbf{f_j}^{(i)}=\mathbf{w}^T_j\mathbf{x}_i({t})$ of linear combinations. 
% Compared to vector autoregression that learns overall time-invariant response space correlations, our \acro{GP} dynamic system approach could learn potential time-varying latent correlation structures through the use of time kernels. 
To extend this, let $\mathbf{f}^{(i)}=[\mathbf{f_{1}}^{(i)},\dots, \mathbf{f_{J}}^{(i)}]^T$ represents the flattened $JT\times 1$ vector consisting of all $J$ tasks. We write $\mathbf{f}^{(i)}$ in a formal multi-task \acro{GP} notation using Kronecker product $\otimes$:
\begin{gather}
\label{eq:multitask}
p(\mathbf{f}^{(i)})\sim\mathcal{GP}\big(\mathbf{0},\mathbf{K}^{(i)}_{\text{task}}\otimes \mathbf{K}^{(i)}_{\text{time}} \big)\\
\label{eq:idikernel}
\mathbf{K}^{(i)}_{\text{task}} = \mathbf{W}_{\text{pop}}^T\mathbf{W}_{\text{pop}} + \mathbf{w}_{{i}}^T\mathbf{w}_{i} + \text{diag}(\mathbf{v})
\end{gather}
where $\mathbf{K}^{(i)}_{\text{task}}$ denotes the unit-individualized task kernel, consisting of the self inner products of population loading $\mathbf{W}_{\text{pop}}=[\mathbf{w}_{1},\dots,\mathbf{w}_{J}]$ for explaining the interpersonal commonality and $J\times 1$ idiographic loading $\mathbf{w}_{{i}}$ for intrapersonal deviations, as well as a task-dependent noise component $\text{diag}(\mathbf{v})=\text{diag}([\sigma_1^2,\dots,\sigma_J^2])$. The Kronecker product $\otimes$ then multiplies each entry in the $J\times J$ task covariance with $\mathbf{K}^{(i)}_{\text{time}}$, and returns the stacked $JT\times JT$ covariance for $\mathbf{f}^{(i)}$. Through the use of time kernel $\mathbf{K}^{(i)}_{\text{time}}$, properties of the latent trait trends such as periodicity or autocorrelation could be incorporated. Here we use the common \acro{RBF} kernel $\mathbf{K}^{(i)}_{\text{time}}(t,t')=\exp\big(-{(t-t')^2}/{\ell_i^2}\big)$ to account for dynamic changes in the latent attributes, whose bandwidth is determined by the unit-specific length scale $\ell_i$, but any other kernel can substitute \acro{RBF} as practitioners see fit. Finally, the latent variables $\mathbf{f^{(i)}}$s are further projected to response space by the ordered logit model.

\subsection{Variational inference}

% inducing no analytical posterior $p(\mathbf{f}^{(i)}\mid \mathbf{y}^{(i)})$ nor marginal likelihood (a.k.a model evidence) $p(\mathbf{y}^{(i)})=\int p(\mathbf{y}^{(i)}\mid \mathbf{f}^{(i)}) p(\mathbf{f}^{(i)})d\mathbf{f}^{(i)}$. Hence

Due to the non-Gaussian ordinal likelihood, we adopt the stochastic variational inference technique (\acro{SVI}) with inducing points introduced in \citep{hensman2015scalable}. Dropping superscript for demonstration, \acro{SVI} utilizes a variational distribution $q(\mathbf{u})=\mathcal{N}(\mu_{\mathbf{u}},\mathbf{\Sigma}_{\mathbf{u}})$ on $m \ll n$ inducing variables $\mathbf{u}$ to approximate $p(\mathbf{f} \mid \mathbf{y})$ using the conditional $p(\mathbf{f} \mid \mathbf{u})$. Hence, the conditional log likelihood $\log p(\mathbf{y}\mid \mathbf{u})$ can be lower bounded by the expected log likelihood w.r.t. $p(\mathbf{f} \mid \mathbf{u})$, after exploiting the non-negativity of Kullback–Leibler (\acro{KL}) divergence between $p(\mathbf{f} \mid \mathbf{u})$ and $p(\mathbf{f} \mid \mathbf{y})$:
\begin{equation}
\label{eq:bound1}
    \log p(\mathbf{y}\mid \mathbf{u})\ge \mathbb{E}_{p(\mathbf{f} \mid \mathbf{u})}\log p(\mathbf{y} \mid \mathbf{f})
\end{equation}
Furthermore, a lower bound on model evidence (\acro{ELBO}) can be obtained by combining Eq. (\ref{eq:bound1}) and an inequality derived by another \acro{KL} divergence $\infdiv{q(\mathbf{u})}{p(\mathbf{u}\mid \mathbf{y})}\ge 0$ (see Appendix \ref{sec:elbo} for details):
\begin{align}
    \log p(\mathbf{y})&\ge  \mathbb{E}_{q(\mathbf{u})}\big[\log p(\mathbf{y}\mid \mathbf{u})\big]-\infdiv{q(\mathbf{u})}{p(\mathbf{u})} \\
  \label{eq:elbo}
    &\ge \mathbb{E}_{q(\mathbf{f})}\big[\log p(\mathbf{y}\mid \mathbf{f})\big]-\infdiv{q(\mathbf{u})}{p(\mathbf{u})}
\end{align}
where the \acro{KL} divergence $\infdiv{q(\mathbf{u})}{p(\mathbf{u})}$ between the variational $q(\mathbf{u})$ and prior $p(\mathbf{u})$ can be computed in closed form as both distributions are Gaussians. The expectation of log likelihood $\log p(\mathbf{y}\mid \mathbf{f})$ under the marginal distribution $q(\mathbf{f})=\int p(\mathbf{f}\mid \mathbf{u}) q(\mathbf{u})d \mathbf{u}$ is intractable but can be numerically approximated using Gauss-Hermite quadrature method. The variational parameters $\mu_{\mathbf{u}}$ and $\mathbf{\Sigma}_{\mathbf{u}}$, individualized loadings $\mathbf{w}_i$ and $\text{diag}(\mathbf{v})$ as well as likelihood parameters $\{b_c\}$s are then optimized to maximize this lower bound. Finally, the predictive likelihood of new $p(\mathbf{y}^*)=\int p(\mathbf{y}^*\mid \mathbf{f}^*)p(\mathbf{f}^*\mid \mathbf{u})q^*(\mathbf{u})  d \mathbf{u}$ is obtained by marginalizing out the optimized $q^*(\mathbf{u})$.

\subsection{Theory testing}
Our \acro{IPGP} framework also naturally facilitates downstream tasks such as domain theory testing between models with and without shared or idiographic components. We adopt Bayes factor, the posterior  $p(\mathcal{M}_i\mid \mathbf{y}) = \frac{ p(\mathbf{y} \mid \mathcal{M}_i)p(\mathcal{M}_i)}{\sum_i p(\mathbf{y} \mid \mathcal{M}_i)p(\mathcal{M}_i)}$ over a pool of models $\{\mathcal{M}_i\}$ conditioning on observation $\mathbf{y}$ with prior weights $p(\mathcal{M}_i)$, as the hypothesis test on whether the latent structures for each individual are indeed distinct or are simply explainable by interpersonal commonality. Specifically, we refer the multi-task model in Eq. (\ref{eq:idikernel}) as the \textit{idiographic} model, and compare it with an \textit{nomothetic} model without unit-specific components: $\mathbf{K}^{\text{pop}}_{\text{task}} = \mathbf{W}_{\text{pop}}\mathbf{W}_{\text{pop}}^T + \text{diag}(\mathbf{v})$.
% \begin{equation}
% \label{eq:popkernel}
% \mathbf{K}^{\text{pop}}_{\text{task}} = \mathbf{W}_{\text{pop}}\mathbf{W}_{\text{pop}}^T + \text{diag}(\mathbf{v})
% \end{equation}

% Consider a pool of models or hypotheses  and its categorical prior distribution $p(\mathcal{M}_i)$, the posterior over this pool of models conditional on observation $\mathbf{y}$ can be computed as:
% \begin{equation}
%     p(\mathcal{M}_i\mid \mathbf{y}) = \frac{ p(\mathbf{y} \mid \mathcal{M}_i)p(\mathcal{M}_i)}{\sum_i p(\mathbf{y} \mid \mathcal{M}_i)p(\mathcal{M}_i)}
% \end{equation}
% When any pair $(\mathcal{M}_i,\mathcal{M}_i)$ has equal prior, their posterior odds ratio reduces to Bayes factor, or ratio of likelihoods: ${p(\mathbf{y}\mid \mathcal{M}_i)}/{p(\mathbf{y}\mid \mathcal{M}_j)}$.

Note that compared to this baseline nomothetic model, our proposed idiographic model in Eq. (\ref{eq:idikernel}) introduces extra unit-level $Jn$ loading parameters that enlarges the optimization space of hyperparameter. Hence, we propose to first learn the interpersonal loading matrix $\mathbf{W}_{\text{pop}}$ using the standard cross-sectional data from a nomothetic model that focuses on learning of population taxonomy, and then use the estimated $\mathbf{W}_{\text{pop}}$ as informative prior in the full model. We will show empirically in Sec. (\ref{sec:experiment}) that with this stronger prior \acro{IPGP} achieves more precise estimation of individual taxonomies.

\section{Experiments} \label{sec:experiment}

We now evaluate \acro{IPGP} in learning idiographic latent taxonomies and predicting actual responses against baseline methods from both psychometrics and Gaussian process literature in three experiments: a simulation study, a re-analysis of a large cross-sectional dataset, and a pilot study of repeated measures of the Big Five \citep{mccrae1992introduction} personality traits. 

%We highlight the predictive ability of \acro{IPGP} through a forecasting and leave-one-trait-out cross validation tasks, and illustrate how \acro{IPGP} better identifies unique  of personality that might advance individualized approaches to psychological diagnosis and inspire new theory.

\subsection{Simulation and ablation}
\paragraph{Setup.} Our simulation considers longitudinal data of $n=10$ units over $T=30$ periods. We assume latent traits of each unit $i$ has dimension $K=5$, and each dimension latent vector is generated independently from a \acro{GP} $\mathbf{x}^{(k)}_{i}(t)\sim\mathcal{GP}(\mathbf{0},\mathbf{K}^{(i)}_\text{time})$ with unit-specific length scale uniformly randomly picked from $\ell^{(i)}_{\text{time}}\in [10,20,30]$. We split $m=20$ batteries into $K$ subsets of size $m/K=4$, such that each subset dominates one dimensional in the latent traits. Specifically, we set high value of $3$ in the population factor loading matrix $\mathbf{W}_{\text{pop}}$ for entries corresponding to the $k$th subset for dimension $k$, and low values drawn from $\text{Unif}[-1,1]$ otherwise. We also set each unit-specific loading $\mathbf{w}_{i}$ from $\text{Unif}[-1,1]$. To introduce sparsity and reverse coding, we randomly set half of the loadings to zero and invert the signs of the remaining half. Finally, we generate the $y_{ijt}$s according to the ordered logit model with $C=5$ levels, and apply $80\%/20\%$ splitting for training and testing.

\begin{table}[H]
\begin{center}
\caption{{\small Comparison of averaged accuracy, log lik and correlation matrix distance between \acro{IPGP} and baselines and ablated models in the simulated study. The full \acro{IPGP} model (indicated in bold) significantly outperforms all ablated and baseline methods in both estimated correlation matrix and either in-sample or out-of-sample prediction in paired-t tests. Results from ablations imply that \acro{IPGP} succeeds in predicting the correct labels due to its idiographic components and proper likelihood, and a well-informed population kernel is crucial in recovering the factor loadings. ``\NA'' indicates baseline software that cannot handle missing values.}}
\label{table:simulation}
\small
\resizebox{\textwidth}{!}{
\begin{tabular}{lrrrrr}
\toprule
 & \multicolumn{1}{c}{\bftab \acro{train acc} $\boldsymbol{\uparrow}$}  & \multicolumn{1}{c}{\bftab \acro{train ll} $\boldsymbol{\uparrow}$} & \multicolumn{1}{c}{\bftab \acro{test acc} $\boldsymbol{\uparrow}$} & \multicolumn{1}{c}{\bftab \acro{test ll} $\boldsymbol{\uparrow}$} & \multicolumn{1}{c}{\bftab \acro{CMD} $\boldsymbol{\downarrow}$}  \\ \midrule
\acro{GRM}  & $0.261 \pm 0.005$ & $-3.556 \pm 0.092$ & $0.261 \pm 0.006$ & $-3.578 \pm 0.098$ & $0.657 \pm 0.021$ \\
\acro{GPCM} & $0.562 \pm 0.017$ & $-2.067 \pm 0.182$ & $0.495 \pm 0.012$ & $-2.409 \pm 0.143$ & $0.545 \pm 0.016$ \\
\acro{SRM}  & $0.286 \pm 0.006$ & $-7.408 \pm 0.063$ & $0.289 \pm 0.008$ & $-7.341 \pm 0.084$ & $0.300 \pm 0.024$ \\
\acro{GPDM}  & \text{0.687 $\pm$ 0.010} & $-4.358 \pm 0.028$ & \text{0.667 $\pm$ 0.010} & $-4.377 \pm 0.029$ & $0.262 \pm 0.016$ \\
\acro{DSEM}  & $0.539 \pm 0.021$ & $-0.961 \pm 0.015$ & \NA & \NA & $0.256 \pm 0.011$ \\
\acro{TVAR}  & $0.554 \pm 0.018$ & $-1.168 \pm 0.014$ & \NA & \NA & $0.987 \pm 0.013$ \\
\midrule
\acro{IPGP-NOM}  & \text{0.807 $\pm$ 0.007} & $-0.535 \pm 0.015$ & \text{ 0.790 $\pm$ 0.008} & $-0.555 \pm 0.017$ & $0.257 \pm 0.009$ \\
\acro{IPGP-IND}  & \text{0.932 $\pm$ 0.003} & $-${ 0.243 $\pm$ 0.008} & \text{0.916 $\pm$ 0.004} & $-$\text{0.267 $\pm$ 0.009} & $0.530 \pm 0.005$ \\
\acro{IPGP-LOW} & \text{0.897 $\pm$ 0.004} & $-0.313 \pm 0.010$ & \text{0.884 $\pm$ 0.005} & $-0.334 \pm 0.011$ & $ 0.397 \pm 0.007$  \\
\acro{IPGP-NP}  & \text{0.898 $\pm$ 0.003} & $-0.318 \pm 0.009$ & \text{0.883 $\pm$ 0.005} & $-0.342 \pm 0.011$ & $0.467 \pm 0.010$  \\
\midrule
\textbf{\acro{IPGP}} & \textbf{0.957 $\pm$ 0.002} & $-$\textbf{0.159 $\pm$ 0.005} & \textbf{0.942 $\pm$ 0.002} & $-$\textbf{0.184 $\pm$ 0.006} & \textbf{0.128 $\pm$ 0.006} \\
\bottomrule
\end{tabular}}
\end{center}
\end{table}

\paragraph{Metrics and baselines.} We consider two sets of metrics for evaluation: (1) the in-sample and out-of-sample predictive accuracy (\acro{ACC}) and log likelihood (\acro{LL}) of the actual responses, (2) the correlation matrix distance (\acro{CMD}) between the estimated factor loading matrix and the true ones, which is defined for two covariance matrices $\mathbf{R}_1,\mathbf{R}_2$ as $\text{d}(\mathbf{R}_1,\mathbf{R}_2)=1-\frac{\text{tr}(\mathbf{R}_1\mathbf{R}_1)}{\norm{\mathbf{R}_1}_f\norm{\mathbf{R}_1}_f}$ \citep{herdin2005correlation} with $l_2$ Frobenius norm. Note that \acro{CMD} becomes zero if $\mathbf{R}_1,\mathbf{R}_2$ are equal up to a scaling factor, and one if they are orthogonal after flattening. We compare \acro{IPGP} to (1) various latent variable models for ordinal responses, including the graded response model (\acro{GRM}) \citep{samejima1969estimation}, the generalized partial credit model (\acro{GPCM}) \citep{muraki1992generalized} and the sequential response model (\acro{SRM}) \citep{tutz1990sequential}, (2) Gaussian process dynamic model (\acro{GPDM}) \citep{damianou2011variational, durichen2014multitask} where the continuous predictions are rounded to the nearest ordinal level, (3) dynamic structural equation model (\acro{DSEM}) \citep{asparouhov2018dynamic, mcneish2023dynamic} with trait-dependent latent variables and (4) time-varying vector autoregression (\acro{TVAR}) with regularized kernel smoothing \citep{haslbeck2021tutorial}. We also compare \acro{IPGP} with several ablated models: (1) \acro{IPGP-NOM} without the idiographic kernel, (2) \acro{IPGP-IND} without the population kernel, (3) \acro{IPGP-LOW} with lower-rank factors of $2$ than actual rank of $5$ in the synthetic setup and (4) \acro{IPGP-NP} where the population kernel is learned from scratch rather than fixed to the informative prior. Note that $\mathbf{W}_{\text{pop}}$ in the full \acro{IPGP} model is fixed as learned from \acro{IPGP-NOM}.

\paragraph{Results.} We use $100$ inducing points and \acro{adam} optimizer of learning rate $0.05$ to optimize \acro{ELBO} for $10$ epoches with batch size of $256$. We repeat our simulation with $25$ different random seeds using 300 Intel Xeon 2680 CPUs. Table \ref{table:simulation} shows comparison of averaged predictive accuracy, log likelihood and correlation matrix distance between \acro{IPGP} and baselines and ablated models in the simulated study. Our \acro{IPGP} model (indicated in bold) significantly outperforms all ablated models and baseline methods in estimated correlation matrix, predictive accuracy and log likelihood of both training and testing sets in paired-t tests. We found that \acro{IPGP} succeeds in predicting the correct labels due to its idiographic components and proper likelihood, since \acro{IPGP-NOM} and \acro{IPGP-GL} are two of the worst ablations for all prediction metrics. In addition, \acro{IPGP-IND} and \acro{IPGP-NP} have the worst correlation matrix estimation, implying that a well-informed population kernel is crucial in recovering the underlying factor structures.

\subsection{Cross-secitonal Factor analysis}\label{sec:loopr}

We next validate the popular Big Five personality theory using standard cross-sectional data via a factor analysis, where a range of factors are tested and then compared according to model evidence. This serves to show that the model works appropriately to detect known latent traits even in non-dynamic settings, and to validate the informative prior for the $\mathbf{W}_{\text{pop}}$ matrix in our next experiment. We utilize an existing dataset called life outcomes of personality replication (\acro{LOOPR}) \citep{soto2019replicable}, which is collected from 5,347 unique participants on the Big Five Inventory \citep{john1999big} consisting of $60$ battery questions. Our validation considers a range of latent trait dimension counts from $K=1,\dots,5$. For each dimension count, we first apply principal component analysis (\acro{PCA}) directly on the correlation matrix of the cross-sectional observations to learn a vanilla population factor loading matrix. We then initialize $\mathbf{W}_{\text{pop}}$ in our model with this vanilla loading matrix, and optimize the loading matrix jointly with the variational parameters. Note that $T=1$ in \acro{LOOPR}, so we drop the idiographic components. 

\begin{table}[H]
\centering
\caption{{\small In-sample accuracy and averaged log lik of our method and baselines for various $K$ in \acro{LOOPR}. Best model for each $K$ is indicated in bold and the best model across different $K$s is further indicated in italic.}}
\label{table:loopr}
\resizebox{\textwidth}{!}{
\begin{tabular}{lrrrrrrrrrr}
\toprule
&  \multicolumn{5}{c}{\bftab  \acro{ACC} $\boldsymbol{\uparrow}$} &   \multicolumn{5}{c}{\bftab  \acro{LL} / \acro{N} $\boldsymbol{\uparrow}$} \\
\cmidrule{2-11}
\bftab \acro{MODEL} & $K=1$ &  $K=2$ &  $K=3$ &  $K=4$ & $K=5$ & $K=1$ &  $K=2$ & $K=3$ & $K=4$ &  $K=5$ \\
\midrule
 \acro{PCA} & 0.106 & 0.099 & 0.123 & 0.217 & 0.192 & $-1.957$ &  $-1.990$ & $-2.009$ & $-2.036$ & $-2.051$ \\
 \acro{GRM} & 0.238 & 0.107 & 0.178 & 0.113 & 0.146  & $-1.838$ &  $-1.832$ & $-1.814$ & $-1.838$ & $-1.841$\\
 \acro{GPCM} & 0.213 & 0.156 & 0.186 & 0.159 & 0.163 & $-1.754$ &  $-1.761$ & $-1.764$ & $-1.750$ & $-1.756$\\
  \acro{SRM} & 0.243 & 0.134 & 0.179 & 0.125 & 0.155 & $-1.784$ &  $-1.784$ & $-1.783$ & $-1.780$ & $-1.767$  \\
  \acro{GPDM} & 0.268 & 0.272 & 0.266 & 0.268 & 0.263  & $-2.155$ &  $-2.158$ & $-2.158$ & $-2.159$ & $-2.158$ \\
 \acro{DSEM} & 0.188 & 0.114 & 0.110 & 0.105 & 0.104  & $-1.997$ &  $-1.960$ & $-1.908$ & $-1.845$ & $-1.775$ \\
 \bftab \acro{IPGP} & \bftab 0.322 & \bftab 0.319 & \textbf{\textit{0.323}} & \bftab 0.318 & \bftab 0.318  & {\bftab $-$1.478} & {\bftab $-$1.477} & {\bftab $-$1.477} & {\bftab $-$1.477} & \textbf{\textit{$-$1.476}}\\
\bottomrule
\end{tabular}}
\end{table}

\begin{figure}[t]
        \begin{subfigure}[b]{0.48\textwidth}
             \includegraphics[width=\linewidth]{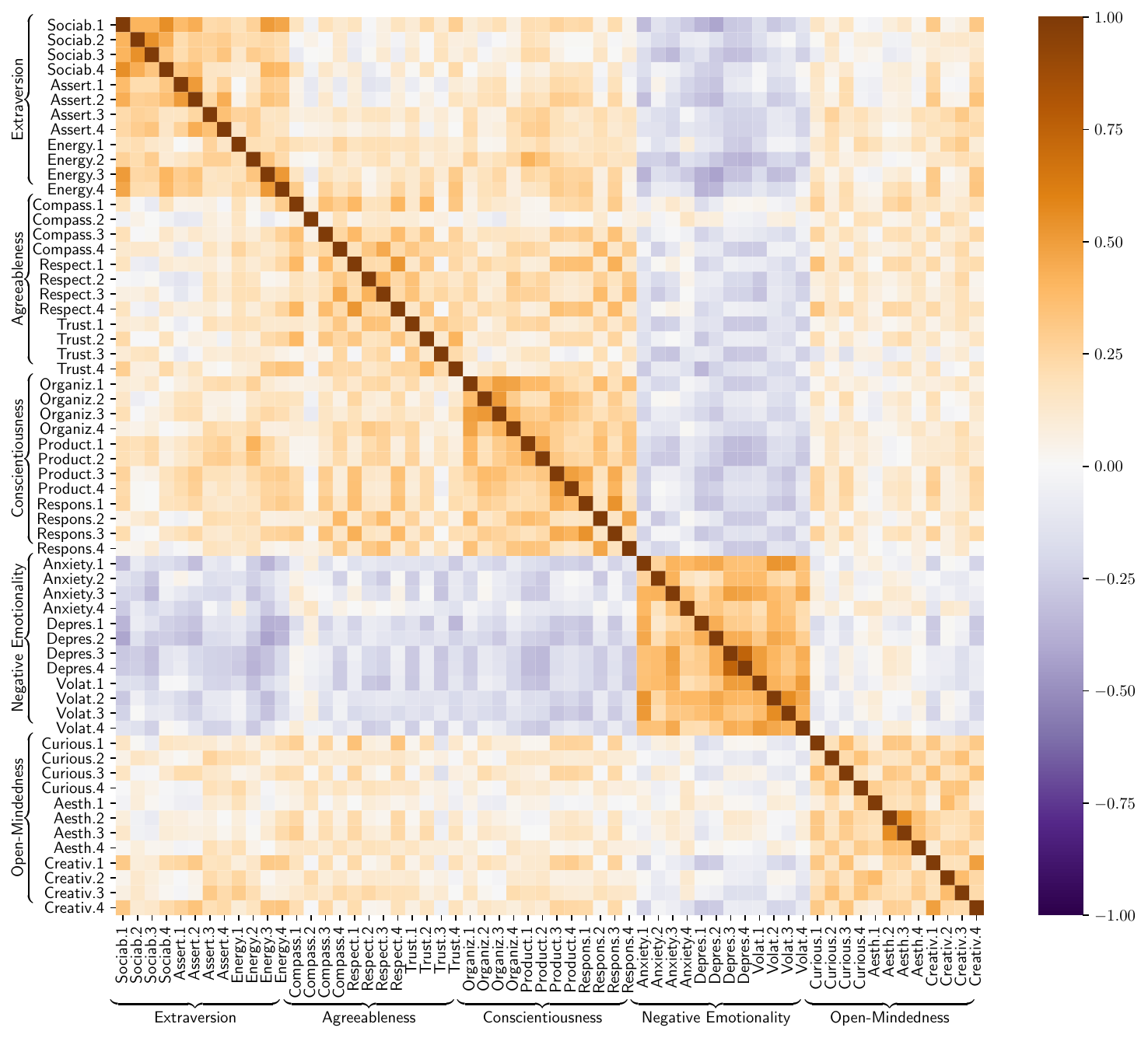}
                % \caption{{\footnotesize Vanilla correlation matrix.}} 
                \label{fig:looprcorr}
        \end{subfigure}\hfill
        \begin{subfigure}[b]{0.48\textwidth}
             \includegraphics[width=\linewidth]{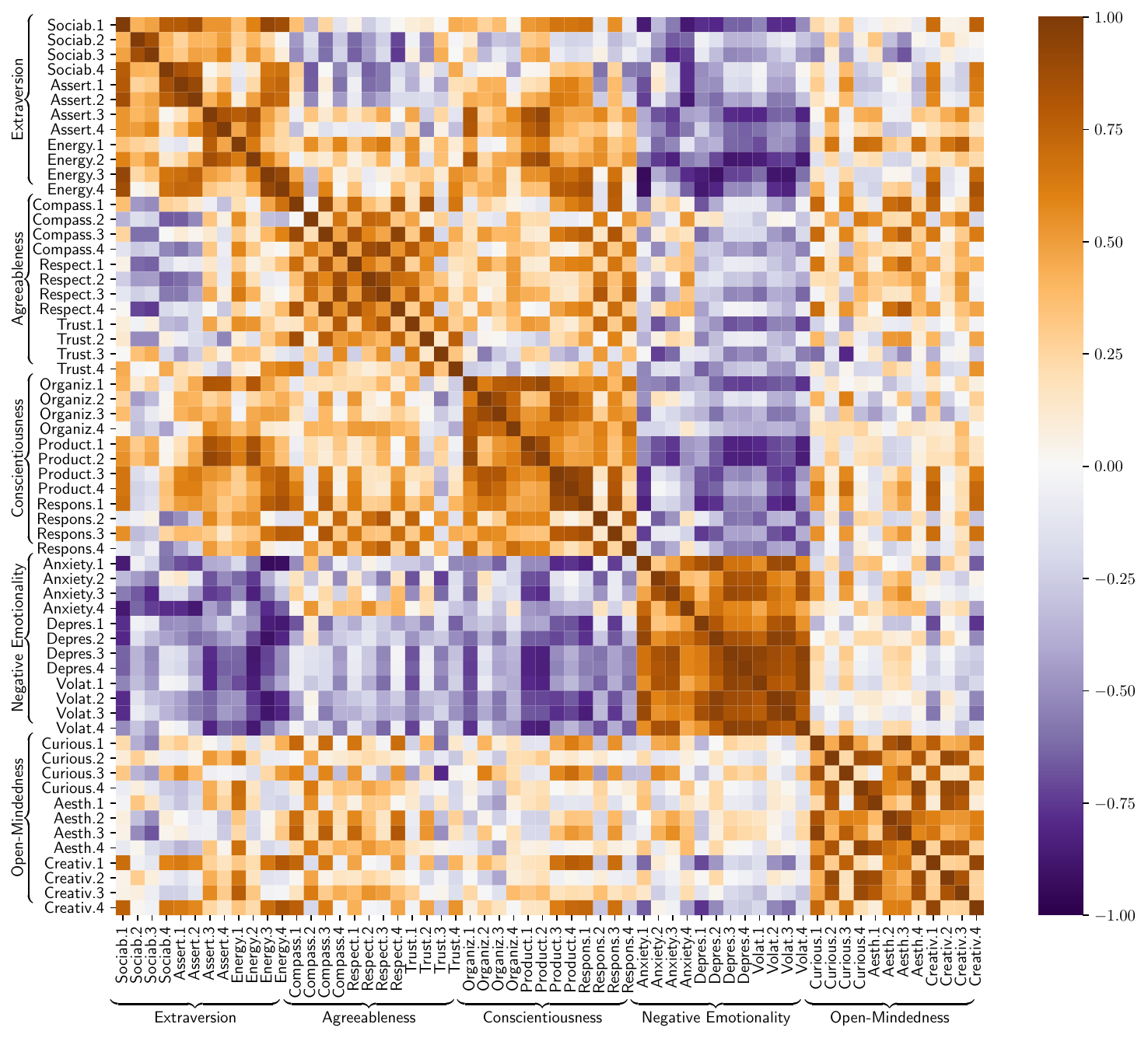}
                % \caption{{\footnotesize Estimated loading matrix under our model.}}
                \label{fig:looprours}
        \end{subfigure}
        \caption{{\small Illustration of raw correlation matrix (left) and our estimated Big Five loading matrix (right). Both correlation matrices displace a \textit{block} pattern, where estimated interpersonal variation show strong correlation between questions within the same factor of the Big Five personalities and weak correlation across different factors. Besides, questions corresponding negative emotionality show minor negative correlation with those corresponding to extraversion and conscientiousness, suggesting trait-by-trait interaction effects.}}
        \label{fig:loopr}
\end{figure}

\paragraph{Validation of Big Five.} Table \ref{table:loopr} shows the predictive accuracy and averaged log likelihood of our method and baseline methods (excluding \acro{TVAR} for lacking low-rank assumption) for various $K$ in \acro{LOOPR}. Best model for each $K$ is indicated in bold numbers and the best model across different $K$s is further indicated in italic numbers. Despite having slightly worse in-sample predictive accuracy than factor $3$ model, \acro{IPGP} with factor $5$ has significant higher model evidence than all the other models, with the second best model is $\exp(-79)$ more unlikely indicated by Bayes factors. Therefore, our results indicate that when psychological measurements are estimated from standard cross-sectional data,  \acro{IPGP} is able to identify the correct factor structure for downstream use.

\paragraph{Estimated interpersonal variation.}
We also show the raw correlation and our estimated Big Five correlation in Figure \ref{fig:loopr}. Both correlation matrices display a \textit{block} pattern, where estimated interpersonal variation show strong correlation between questions within the same factor of the Big Five and weak correlation across different factors. In addition, questions corresponding negative emotionality show minor negative correlation with those corresponding to extraversion and conscientiousness, suggesting appropriate trait-by-trait interaction effects.

\subsection{Longnitudinal Pilot Study}\label{sec:longitudinal}

To further demonstrate \acro{IPGP} in longitudinal setting for learning idiographic psychological taxonomies, we collected an intensive longitudinal data using experience sampling measures (\acro{ESM}). We  highlight the predictive ability of \acro{IPGP} through a forecasting and a leave-one-trait-out cross validation task, and illustrate how \acro{IPGP} identifies unique taxonomies of personality that might advance individualized approaches to psychological diagnosis and inspire new theory.

\paragraph{Data collection.} In \acro{ESM} design, each participant was asked to complete personality assessments six times per day for three weeks, resulting maximum $126$ assessments per person. With $93$ valid student participants, we acquired 8,770 assessments in total with an average of $94$ assessments per person. The personality assessment is derived from the \acro{BFI-2} \citep{soto2017next} to ensure identification of latent factors and ample coverage of the late trait space. The \acro{BFI-2} includes $60$ items with four unique items assessing each of the three different sub-factors for each Big-Five domains. We removed one item for each sub-factors that are not appropriate for contextualized assessments of \acro{ESM} design. To mitigate the fatigue and learning effect from repeated measures, we employed a planned missing design where participants were randomly tested on only two out of three items assessing the same sub-factors, resulting only 30 items for each assessment.

%  \begin{table}[H]
%   \caption{{\small In-sample prediction and averaged log likelihood of our proposed model (\acro{IPGP}) and baselines for the longitudinal data, as well as Bayes factors to \acro{IPGP}. ``\NA'' indicates self comparison of Bayes factors.\\}}
%     \label{table:longitudinal}
% \centering
% \begin{tabular}{lrrc}
% \toprule
%   & \bftab \acro{ACC} & \bftab \acro{LL}/\acro{N} & \bftab $\log$(\acro{BF}) \\
% \midrule
% \acro{GRM} & 0.210 & $-2.266$ & $\num{-2.32e5}$  \\
% \acro{GPCM} & 0.288 & $-1.516$ & $\num{-3.80e4}$  \\
% \acro{SRM} & 0.260 & $-1.927$ & $\num{-1.44e5}$  \\
% \acro{GPDM} & 0.382 & $-3.865$ & $\num{-7.80e5}$ \\
% \acro{DSEM} & 0.226 & $-1.399$ & $\num{-7.72e3}$  \\
% \acro{TVAR} & 0.382 & $-1.546$ & $\num{-4.47e4}$  \\
%  \acro{IPGP-NOM} & 0.403 & $-1.410$ &  $\num{-1.06e4}$  \\
% \bftab \acro{IPGP} & \bftab 0.417 & {\bftab $-$1.369} & \multicolumn{1}{c}{\NA}  \\
% \bottomrule
% \end{tabular}
% \end{table}

\paragraph{Comparison between nomothetic and idiographic models.} We run the full \acro{IPGP} model with idiographic component and unit-specific time kernel on the collected longitudinal data. Again we set the ranks of the population and individual loading matrices to $5$ and $1$ respectively, and incorporate the prior knowledge of the cross-sectional data by fixing the population loadings as the Big Five loadings estimated in Sec. (\ref{sec:loopr}) and optimizing the individual loadings. We contrast our proposed idiographic model (\acro{IPGP}) and baselines in Table \ref{table:longitudinal}, which shows the in-sample prediction, averaged log likelihood and Bayes factors. We found that \acro{IPGP} outperforms \acro{IPGP-NOM} with higher predictive accuracy and log likelihood, and is favored decisively by a Bayes factor of $\exp(\num{1.06e4})$.

\begin{minipage}{\textwidth}
  \begin{minipage}[b]{0.48\textwidth}
      \begin{table}[H]
  \captionof{table}{{\small In-sample prediction and averaged log likelihood of our proposed model (\acro{IPGP}) and baselines for the longitudinal data, as well as Bayes factors to \acro{IPGP}. ``\NA'' indicates self comparison of Bayes factors.\\}}
    \label{table:longitudinal}
\begin{tabular}{lrrc}
\toprule
  & \bftab \acro{ACC} & \bftab \acro{LL}/\acro{N} & \bftab $\log$(\acro{BF}) \\
\midrule
\acro{GRM} & 0.210 & $-2.266$ & $\num{-2.32e5}$  \\
\acro{GPCM} & 0.288 & $-1.516$ & $\num{-3.80e4}$  \\
\acro{SRM} & 0.260 & $-1.927$ & $\num{-1.44e5}$  \\
\acro{GPDM} & 0.382 & $-3.865$ & $\num{-7.80e5}$ \\
\acro{DSEM} & 0.226 & $-1.399$ & $\num{-7.72e3}$  \\
\acro{TVAR} & 0.382 & $-1.546$ & $\num{-4.47e4}$  \\
 \acro{IPGP-NOM} & 0.403 & $-1.410$ &  $\num{-1.06e4}$  \\
\bftab \acro{IPGP} & \bftab 0.417 & {\bftab $-$1.369} & \multicolumn{1}{c}{\NA}  \\
\bottomrule
\end{tabular}
\end{table}
  \end{minipage}
  \hfill
  \begin{minipage}[b]{0.48\textwidth}
  \begin{figure}[H]
    \centering
     \includegraphics[width=1\linewidth, height=0.7\linewidth]{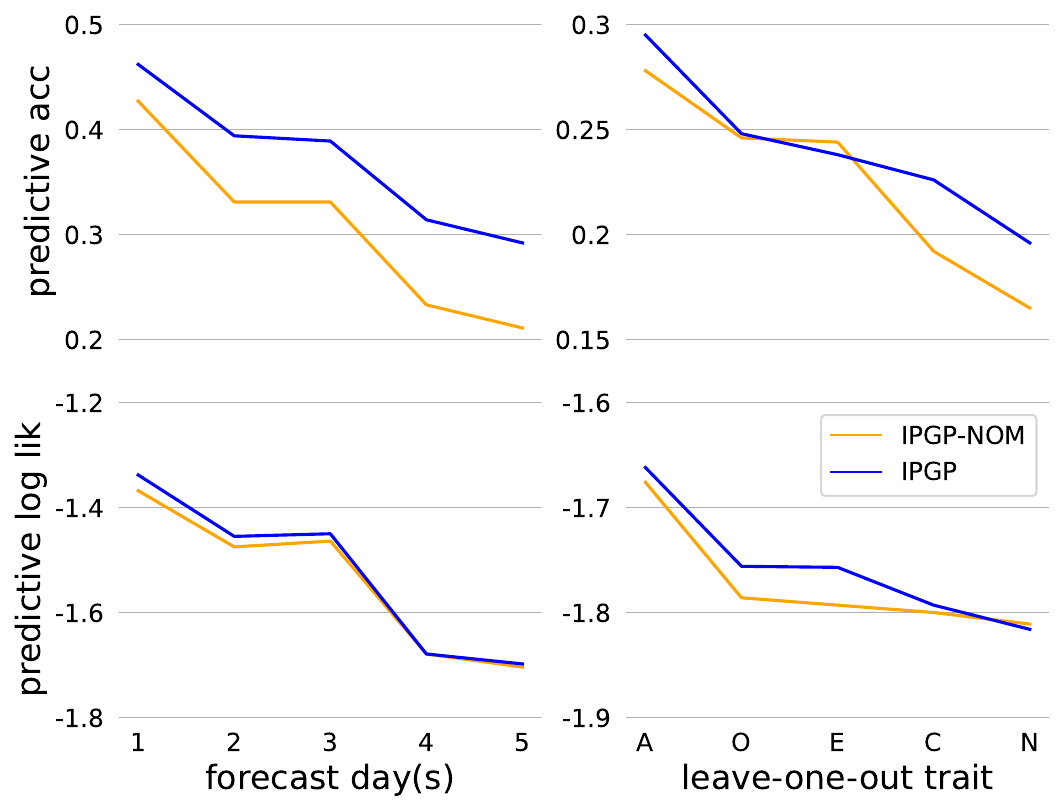}
    \captionof{figure}{Predictive accuracy and log lik of \acro{IPGP} and \acro{IPGP-NOM} for the forecasting task and leave-one-trait-out cross-validation task.}
    \label{fig:longitudinalpred}
    \end{figure}
    \end{minipage}
\end{minipage}

% \begin{figure}[H]
%     \centering
%      \includegraphics[width=0.6\linewidth, height=0.4\linewidth]{Figures/last_acc_ll.pdf}
%     \captionof{figure}{Predictive acc and log lik of \acro{IPGP} and \acro{IPGP-NOM} for the forecasting task and leave-one-trait-out cross-validation task.}
%     \label{fig:longitudinalpred}
% \end{figure}

\paragraph{Predictive performance of \acro{IPGP}.}
We also evaluate the out-of-sample performance of the idiographic and nomothetic models using two prediction tasks: forecasting future responses and leave-one-trait-out cross validation. For the forecasting task, we train both models with data from the first $40$ days and predict future responses for the last $5$ days. For the cross validation task, we predict responses of each  trait by training on data belonging to the other four traits. Figure (\ref{fig:longitudinalpred}) shows the predictive accuracy and log likelihood of \acro{IPGP} and \acro{IPGP-NOM} for the forecasting task over varying horizons and the leave-one-trait-out cross-validation task. \acro{IPGP} has consistently better performance than \acro{IPGP-NOM} in both tasks except for being slightly less accurate in predicting extraversion. Overall, \acro{IPGP} is favored than \acro{IPGP-NOM} by Bayes factors of $\exp(43)$ and $\exp(716)$ in these tasks.

\begin{figure}[H]
        \begin{subfigure}[b]{0.25\textwidth}
             \includegraphics[width=\linewidth]{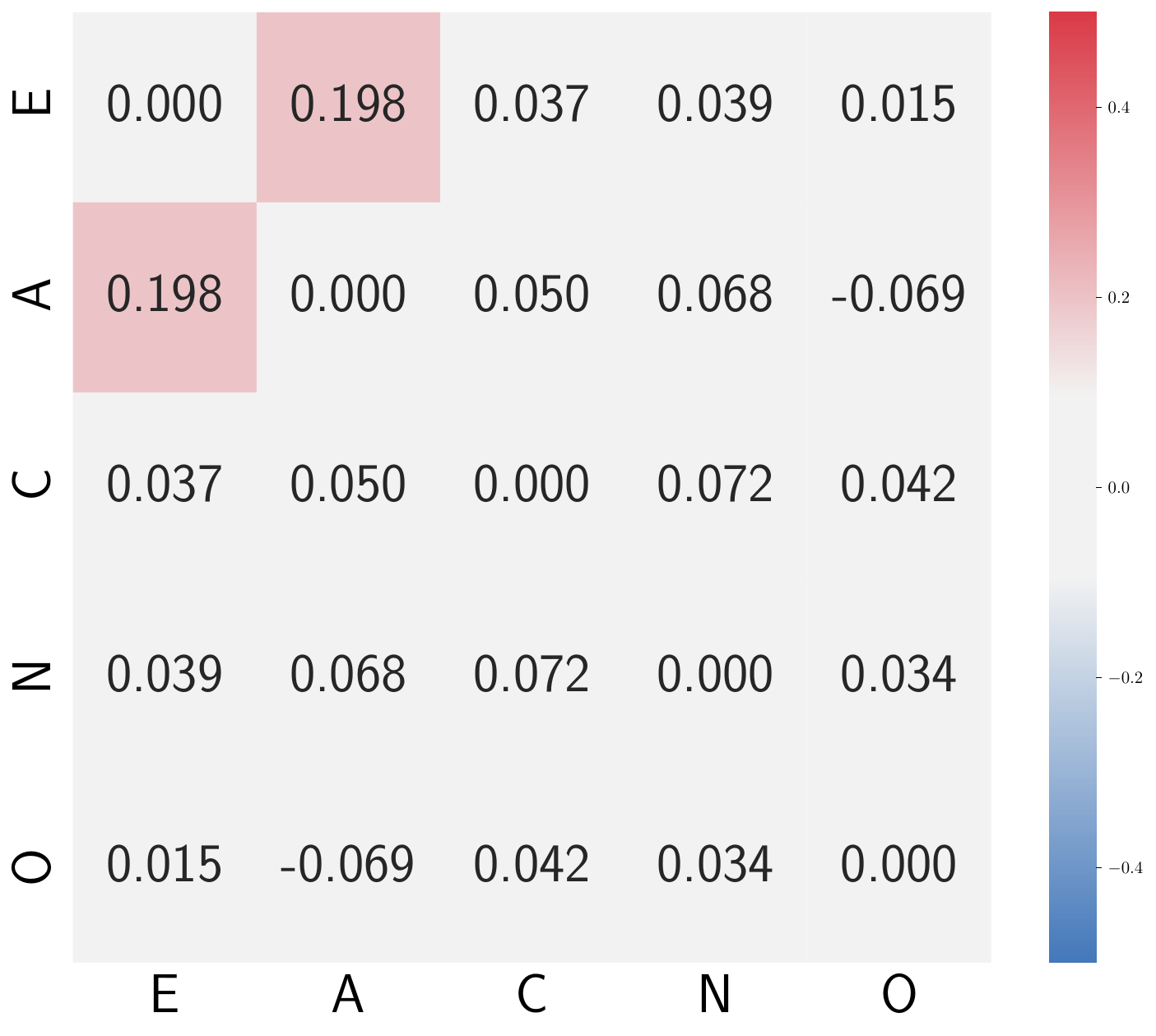}
                \label{fig:longitudinal1}
        \end{subfigure}\hfill
        \begin{subfigure}[b]{0.25\textwidth}
             \includegraphics[width=\linewidth]{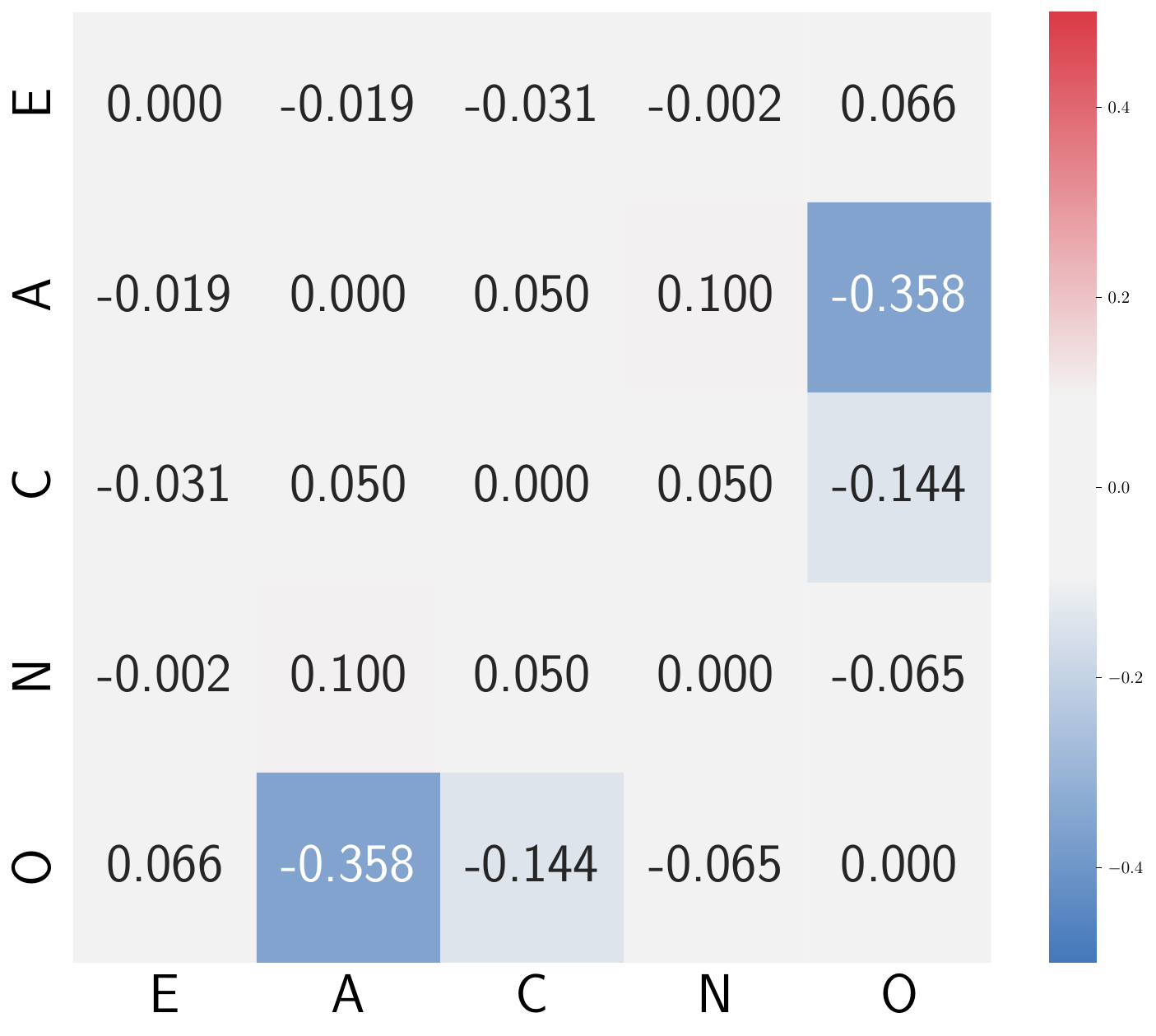}
                \label{fig:longitudinal2}
        \end{subfigure}\hfill
        \begin{subfigure}[b]{0.25\textwidth}
             \includegraphics[width=\linewidth]{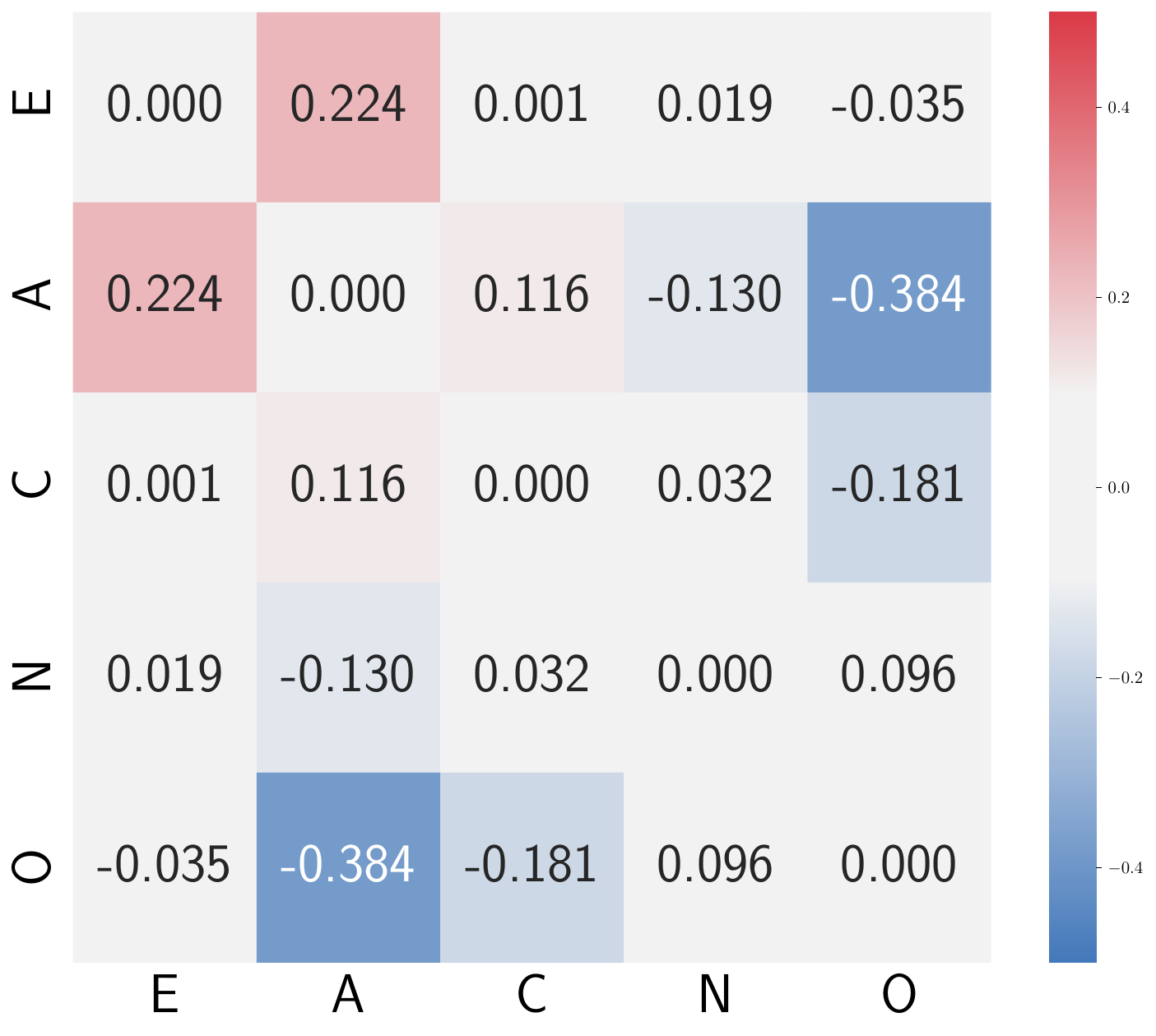}
                \label{fig:longitudinal3}
        \end{subfigure}\hfill
        \begin{subfigure}[b]{0.25\textwidth}
             \includegraphics[width=\linewidth]{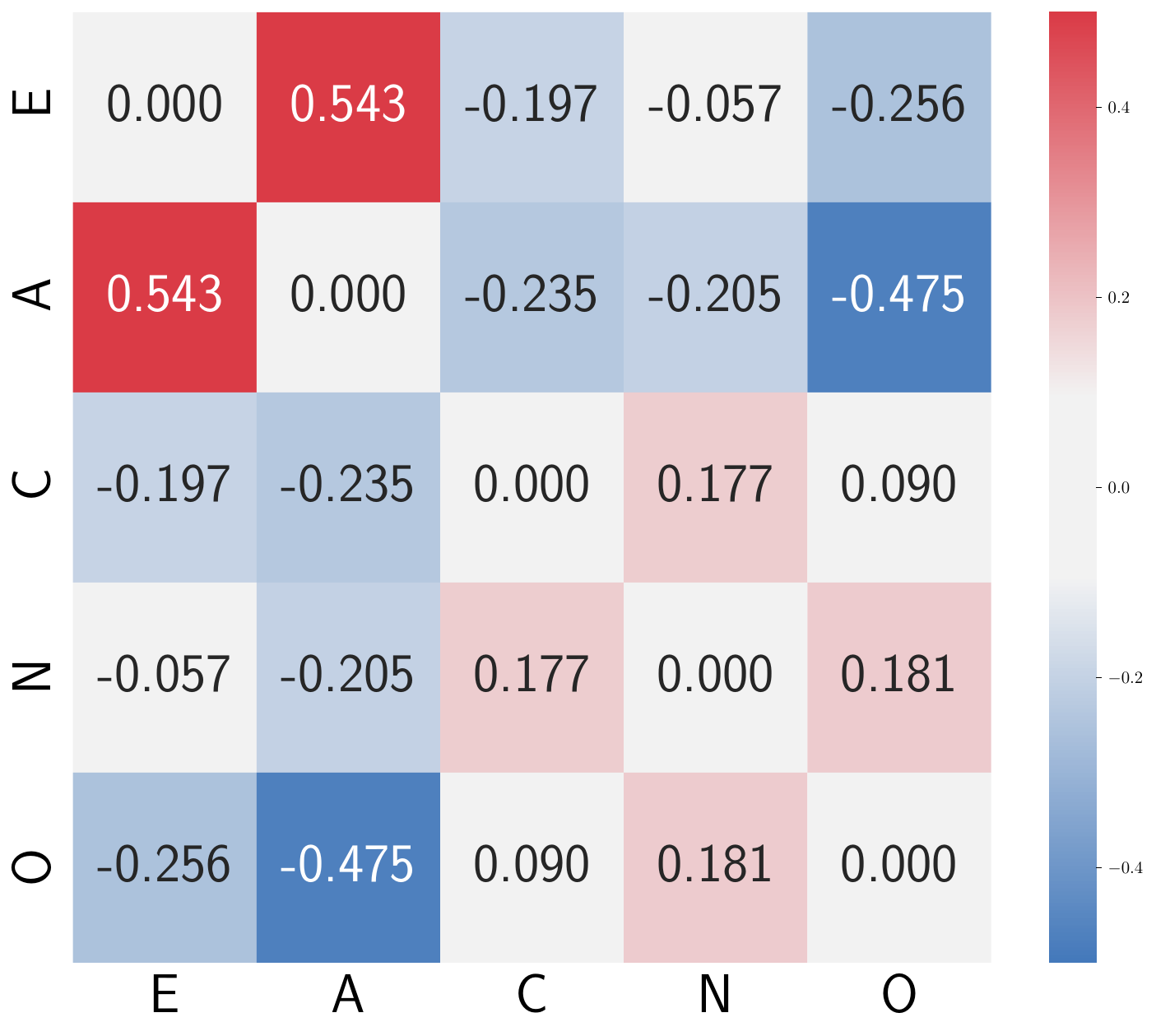}
                \label{fig:longitudinal4}
        \end{subfigure}
        \caption{{\small Four residual correlations as identified by our k-mean clustering. Each heatmap displays the trait-level residual correlation averaged across corresponding batteries for one cluster, with darker red and blue indicating larger positive and negative deviations. For instance, agreeableness (A) is more correlated to extraversion (E) than the population profile in the first profile, but less correlated to openness (O) in the second profile. Moreover, these two directions of deviations are even exacerbated in the third and fourth profiles.}}
        \label{fig:longitudinal}
\end{figure}

\paragraph{Discovery of unique taxonomies.} Despite our small cohort size ($93$ respondents), we also manage to identify distinct profiles of personality that substantially differ from the interpersonal commonality. Specifically, we first perform a k-mean clustering using all $93$ estimated individual correlation matrix with \acro{CMD} as the distance metric, and then compute the residual correlation between each estimated clustering centroid and the population correlation. Figure (\ref{fig:longitudinal}) illustrates four residual correlations as identified by our $k$-mean clustering. Each heatmap displays the trait-level residual correlation averaged across corresponding batteries for one cluster, with darker red and blue indicating larger positive and negative deviations. For instance, agreeableness (A) is more correlated to extraversion (E) than population profile in the first profile, but less correlated to openness (O) in the second profile. Moreover, these two directions of deviations are even exacerbated in the third and fourth profiles. 

The findings of unique taxonomies of personality profiles also suggest a potential solution to the important idiographic vs. nomothetic debate in personality science and psychometric sciences. Considering the limitations of both the idiographic approach and the nomothetic approach, the current findings suggest that the ideal solution lies somewhere in between completely idiographic and nomothetic. The four distinct profiles derived from Big-Five accommodated individuals’ uniqueness by suggesting people could deviate from a common taxonomy, which offers us meaningful insights into their unique motivations, behavioral patterns, and self-concepts. For example, the large overlap between Extraversion and Agreeableness for individuals who fit Profile 4 may arise because they tend to act warm in social and outgoing manner (e.g., the person at a party that engages and be nice to everyone). Meanwhile, these distinct profiles will help to improve the model predictability from the $N=1$ models by learning from individuals with similar profiles.

\section{Related work}\label{sec:related}

\textbf{Idiographic assessment} emphasizes the important aspects of individuals otherwise missing in oversimplified taxonomies of psychological behaviors \citep{hamaker2009idiographic}. Empirical evidence across many psychometrics fields has shown the lack of generalizability of the nomothetic models only focusing on interpersonal variation \citep{molenaar2004manifesto}. Hence, \citet{song2012bayesian} incorporated simple random effect method with dynamic factor models for analyzing psychological processes.
\citet{jongerling2015multilevel} proposed a multilevel first-order autoregressive model with random intercepts to measure daily positive effects over several weeks.
\citet{beltz2016bridging} combined the nomothetic and idiographic approaches in analyzing clinical data by adding individual components to the group iterative multiple model (\acro{GIMME}). However, all of these methods focus on modeling in response space rather than latent space for ordinal survey data. 

\textbf{Gaussian process latent variable model} (\acro{GPLVM}) is a dimensional reduction method for Gaussian data, where the latent variables are optimized after integrating out the function mappings \citep{lawrence2003gaussian, pmlr-v151-lalchand22a}. Our proposed framework differs from \acro{GPLVM} as we optimize the factor loading matrix while marginalizing the latent variables. In addition, our model contrasts \acro{GPLVM} and (variational) Gaussian Process dynamical model (\acro{GPDM}) \citep{NIPS2005_ccd45007, damianou2011variational} in our non-Gaussian ordered logistic observation model. Finally, our longitudinal framework with stochastic variational inference learning differs from the static \acro{GP} item response theory (\acro{GPIRT}) \citep{duck2020gpirt} with more computationally demanding Gibbs sampling.

 % Compared to existing dynamic measurement models such as dynamic item response theory \citep{rijmen2003nonlinear, reise2009item, dumas2020dynamic}, longitudinal structural equations \citep{asparouhov2018dynamic, mcneish2023dynamic} and vectorized autoregression \citep{haslbeck2021tutorial}, we show that our framework allows more precise individualization from a common (shared) taxonomy, better extrapolations of future responses from past as well as one trait from the others, and provides a more flexible way to examine whether individuals are unique in their latent structures.

\textbf{Longitudinal measurement models} integrate temporal dynamics into psychological theories with growing popularity of longitudinal design in survey methods \citep{jebb2015time, ariens2020time}. For instance, families of longitudinal structural equation models (\acro{SEM}) such as multiple-group longitudinal \acro{SEM} and longitudinal growth curve model were developed for repeated measurement studies \citep{little2013longitudinal}, where M\textit{plus} software was developed later for dynamic \acro{SEM} with Bayesian Gibbs sampling \citep{asparouhov2018dynamic, mcneish2023dynamic}. Dynamic item response models \citep{rijmen2003nonlinear,reise2009item, dumas2020dynamic} and time-varying vector autoregressive model \citep{lu2018bayesian, haslbeck2021tutorial} were also proposed to estimate the trajectories of latent traits. Despite previous work in behavioral literature focusing on Gaussian observations \citep{durichen2014multitask}, multi-task Gaussian process time series has not yet been exploited for survey experiments with non-Gaussian likelihood when exact inference is not plausible.

\section{Conclusion} \label{sec:conclusion}

We propose a novel idiographic personality Gaussian process (\acro{IPGP}) model for personalized psychological assessment and learning of intrapersonal taxonomy from longitudinal ordinal survey data, an under-explored setup in Gaussian process dynamic system literature. We exploit Gaussian process coregionalization for capturing between-battery structure and stochastic variational inference for scalable inference. Future directions include adaptation of \acro{IPGP} to other psychological studies such as emotion, and incorporation of contextual information such as behaviors or current activities. 

Our proposed \acro{IPGP} framework also provides insights to domain theory testing, addressing the substantive debate in psychometrics surrounding the shared versus unique structures of psychological features.  Our experimental results show that \acro{IPGP} is decisively favored than the nomothetic baseline, and substantive deviations from the common trend persist in considerable individuals. Hence, our framework has a great potential in advancing individualized approaches to psychological diagnosis.

\begin{ack}
This work was supported by the 2023 Seed Grant of Transdisciplinary Institute in Applied Data Sciences at Washington University.

\end{ack}

% References
\bibliography{reference}
\bibliographystyle{unsrtnat}

\newpage
\appendix

\section{Mathematical Details of Evidence Lower Bound}\label{sec:elbo}
We provide the full mathematical details of the evidence lower bound defined in Eq. (\ref{eq:elbo}). As \acro{KL} divergence is always non-negative, we first consider the \acro{KL} divergence between $p(\mathbf{f}\mid \mathbf{u})$ and $p(\mathbf{f} \mid \mathbf{y})$:
\begin{align}
    \infdiv{p(\mathbf{f} \mid \mathbf{u})}{p(\mathbf{f} \mid \mathbf{y})} &=\mathbb{E}_{p(\mathbf{f} \mid \mathbf{u})}\log \frac{p(\mathbf{f} \mid \mathbf{u})}{p(\mathbf{f} \mid \mathbf{y})} \\
    &=\mathbb{E}_{p(\mathbf{f} \mid \mathbf{u})}\log \frac{p(\mathbf{f} \mid \mathbf{u})p(\mathbf{y})}{p(\mathbf{y} \mid \mathbf{f})p(\mathbf{f})}\\
    &=\mathbb{E}_{p(\mathbf{f} \mid \mathbf{u})}\log \frac{p(\mathbf{f} \mid \mathbf{u})p(\mathbf{y} \mid \mathbf{u}) p(\mathbf{u})}{p(\mathbf{y} \mid \mathbf{f})p(\mathbf{f})}\\
    &=\mathbb{E}_{p(\mathbf{f} \mid \mathbf{u})}\log \frac{p(\mathbf{y} \mid \mathbf{u})}{p(\mathbf{y} \mid \mathbf{f})}\\
    &=\log p(\mathbf{y} \mid \mathbf{u})-\mathbb{E}_{p(\mathbf{f} \mid \mathbf{u})}\log {p(\mathbf{y} \mid \mathbf{f})} \ge 0
\end{align}
Moving $\mathbb{E}_{p(\mathbf{f} \mid \mathbf{u})}\log {p(\mathbf{y} \mid \mathbf{f})} $ to the R.H.S of the above inequality will lead to Eq. (\ref{eq:bound1}). We then exploit the inequality given by $\infdiv{q(\mathbf{u})}{p(\mathbf{u}\mid \mathbf{y})}\ge 0$:
\begin{align}
    \infdiv{q(\mathbf{u})}{p(\mathbf{u}\mid \mathbf{y})} &=\mathbb{E}_{q(\mathbf{u})}\log \frac{q(\mathbf{u})}{p(\mathbf{u}\mid \mathbf{y})}\\
    &=\mathbb{E}_{q(\mathbf{u})}\log \frac{q(\mathbf{u}) p(\mathbf{y})}{p(\mathbf{y}\mid \mathbf{u})p(\mathbf{u})}\\
    &= - \mathbb{E}_{q(\mathbf{u})}\log p(\mathbf{y}\mid \mathbf{u}) + \infdiv{q(\mathbf{u})}{p(\mathbf{u})} + \log p(\mathbf{y}) \ge 0
\end{align}
Rearranging the above inequality, applying Eq. (\ref{eq:bound1}) and  exploiting notation $q(\mathbf{f})=\int p(\mathbf{f}\mid \mathbf{u}) q(\mathbf{u})d \mathbf{u}$ leads to the \acro{ELBO}:
\begin{align}
   \log p(\mathbf{y}) &\ge \mathbb{E}_{q(\mathbf{u})}\log p(\mathbf{y}\mid \mathbf{u}) - \infdiv{q(\mathbf{u})}{p(\mathbf{u})}\\
&=\mathbb{E}_{q(\mathbf{u})}\big[\mathbb{E}_{p(\mathbf{f} \mid \mathbf{u})}\log {p(\mathbf{y} \mid \mathbf{f})} \big] - \infdiv{q(\mathbf{u})}{p(\mathbf{u})}\\
&=\mathbb{E}_{q(\mathbf{f})}\log {p(\mathbf{y} \mid \mathbf{f})} - \infdiv{q(\mathbf{u})}{p(\mathbf{u})}
\end{align}

\newpage
\section{Estimated Correlations of Selective Individuals}

\begin{figure}[H]
        \begin{subfigure}[b]{0.5\textwidth}
             \includegraphics[width=\linewidth]{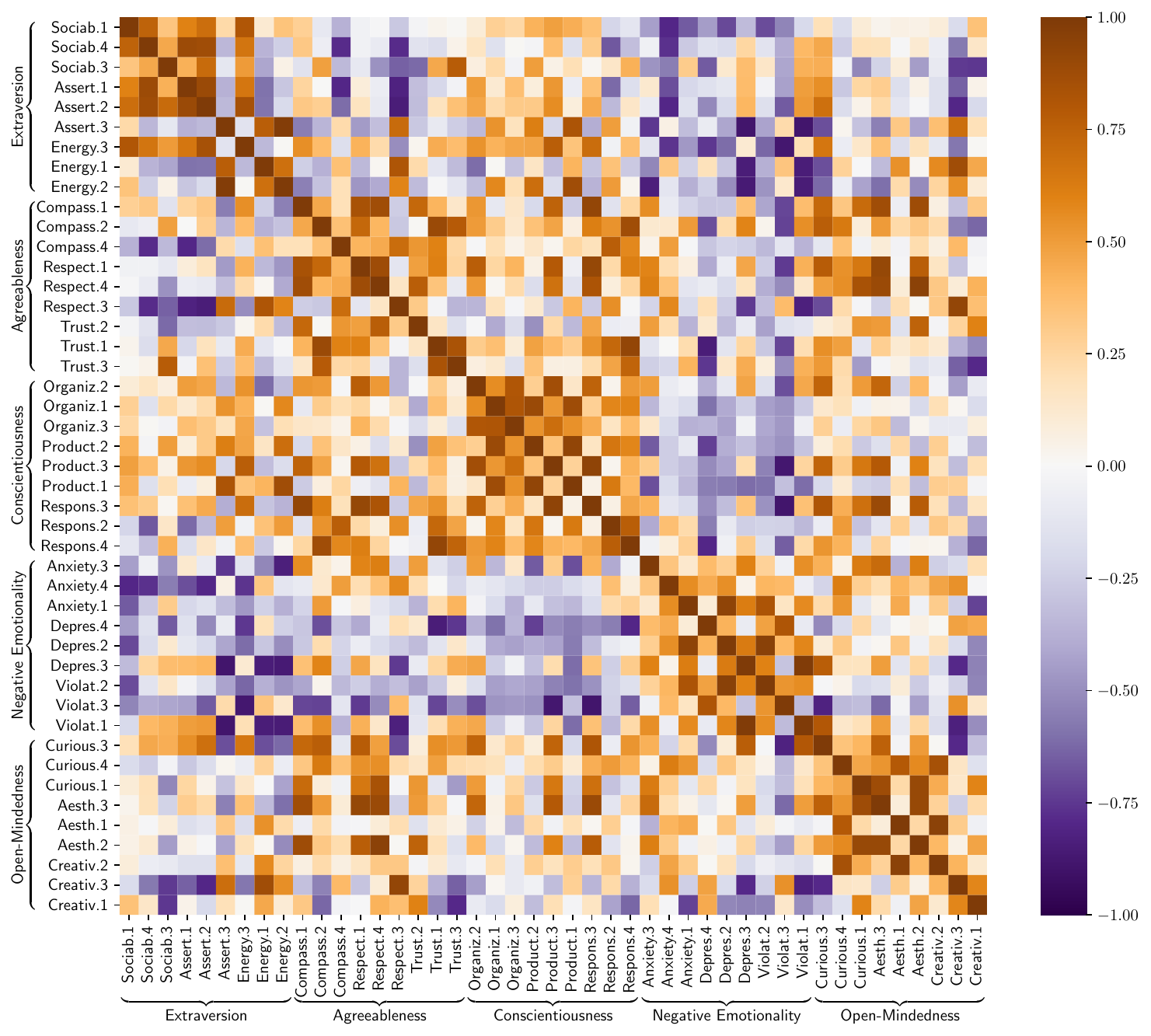}
        \end{subfigure}\hspace{0.2cm}
        \begin{subfigure}[b]{0.5\textwidth}
             \includegraphics[width=\linewidth]{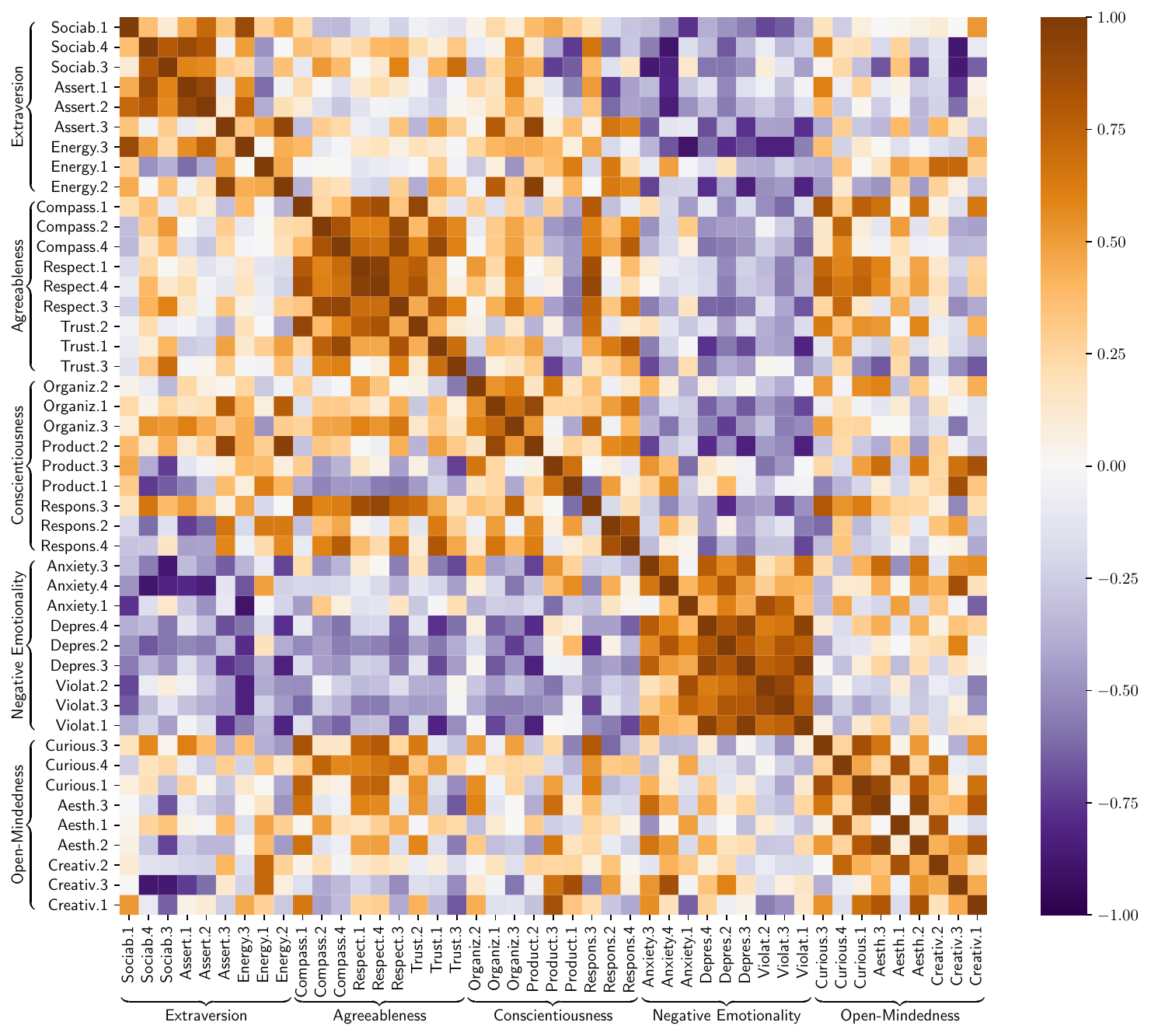}
        \end{subfigure}\\
        \begin{subfigure}[b]{0.5\textwidth}
             \includegraphics[width=\linewidth]{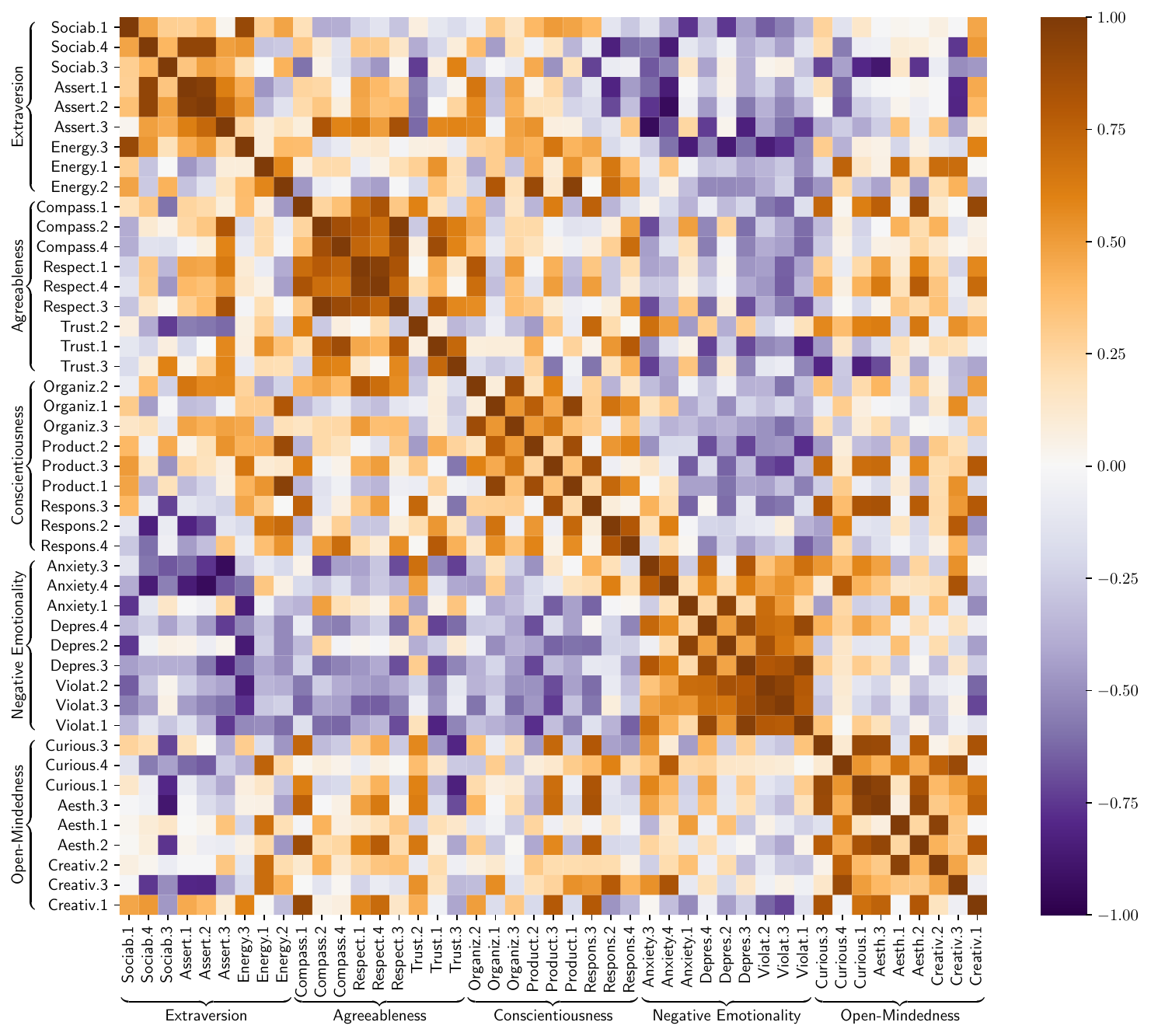}
        \end{subfigure}\hspace{0.2cm}
        \begin{subfigure}[b]{0.5\textwidth}
             \includegraphics[width=\linewidth]{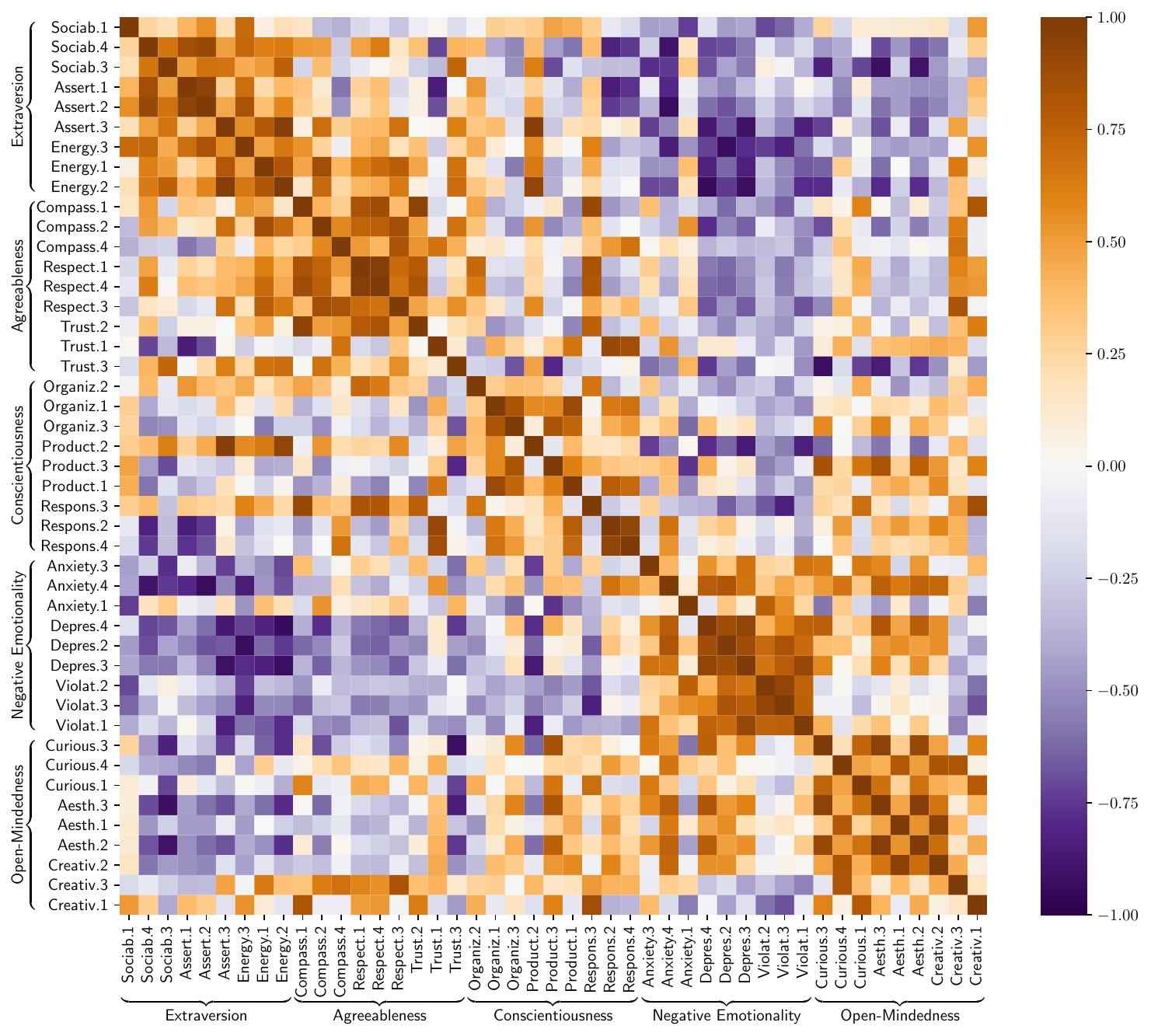}
        \end{subfigure}
        \caption{{\small Estimated correlations of selective individuals for the identified four profiles in the longitudinal study.}}
        \label{fig:correlationprofile}
\end{figure}

Figure (\ref{fig:correlationprofile}) shows the estimated correlations of selective individuals for the identified four profiles in the longitudinal study.

\end{document}